\documentclass[10pt,twocolumn,letterpaper]{article}

\usepackage[]{cvpr}   
\usepackage{multirow}
\usepackage{float}
\usepackage{textcomp}

\usepackage{pifont}          
\usepackage{graphicx}
\usepackage{booktabs}
\usepackage{siunitx}
\usepackage{makecell} 
\usepackage{amsmath} 
\usepackage{xcolor}    % \color

\newcommand\blfootnote[1]{%
  \begingroup
  \renewcommand\thefootnote{}\footnote{#1}%
  \addtocounter{footnote}{-1}%
  \endgroup
}

\newcommand{\cmark}{\ding{51}}  % ✓
\newcommand{\xmark}{\ding{55}}  % ✗
\newcommand{\xovercmark}{%
  \ooalign{%
    \hidewidth\raisebox{0.1ex}{\xmark}\hidewidth\cr
    \cmark\cr
  }%
}

\definecolor{cvprblue}{rgb}{0.21,0.49,0.74}
\usepackage[pagebackref,breaklinks,colorlinks,allcolors=cvprblue]{hyperref}

\def\paperID{***} % *** Enter the Paper ID here
\def\confName{CVPR}
\def\confYear{2026}

\title{UniDex: A Robot Foundation Suite for Universal Dexterous Hand Control from Egocentric Human Videos}

\author{Gu Zhang$^{1,2,\ast\dagger}$, 
Qicheng Xu$^{1,\ast}$, 
Haozhe Zhang$^{1,\ast}$,
Jianhan Ma$^{2,\ast}$,
Long He$^{1,\ast}$,
Yiming Bao$^{1,\ast}$,
\\
Zeyu Ping$^{3}$,
Zhecheng Yuan$^{1,2}$,
Chenhao Lu$^{1}$,
Chengbo Yuan$^{1}$,
Tianhai Liang$^{1}$,
Xiaoyu Tian$^{1}$,\\
Maanping Shao$^{1}$,
Feihong Zhang$^{1}$,
Mingyu Ding$^{4}$,
Yang Gao$^{1,2}$,
Hao Zhao$^{1}$,
Hang Zhao$^{1,2}$,
Huazhe Xu$^{1,2}$\\
[2mm]
$^1$~Tsinghua University \quad
$^2$~Shanghai Qizhi Institute \\
$^3$~Sun Yat-sen University \quad
$^4$~The University of North Carolina at Chapel Hill
\\ 
\normalsize{
$^\ast$Core Contributors \quad$^\dagger$Project Lead     }
\\
\normalsize{
\url{https://unidex-ai.github.io/}
}
}

\begin{document}

\maketitle

\blfootnote{* Core Contributors: see contribution list \hyperref[sec:contrib]{here}.}

\begin{abstract}
Dexterous manipulation remains challenging due to the cost of collecting real-robot teleoperation data, the heterogeneity of hand embodiments, and the high dimensionality of control. We present UniDex, a robot foundation suite that couples a large-scale robot-centric dataset with a unified vision–language–action (VLA) policy and a practical human-data capture setup for universal dexterous hand control. \textbf{First}, we construct UniDex-Dataset, a robot-centric dataset over 50K trajectories across eight dexterous hands (6–24 DoFs), derived from egocentric human video datasets. To transform human data into robot-executable trajectories, we employ a human-in-the-loop retargeting procedure to align fingertip trajectories while preserving plausible hand–object contacts, and we operate on explicit 3D pointclouds with human hands masked to narrow kinematic and visual gaps. \textbf{Second}, we introduce the Function–Actuator–Aligned Space (FAAS), a unified action space that maps functionally similar actuators to shared coordinates, enabling cross-hand transfer. Leveraging FAAS as the action parameterization, we train UniDex-VLA, a 3D VLA policy pretrained on UniDex-Dataset and finetuned with task demonstrations. \textbf{In addition}, we build UniDex-Cap, a simple portable capture setup that records synchronized RGB-D streams and human hand poses and converts them into robot-executable trajectories to enable human–robot data co-training that reduces reliance on costly robot demonstrations. On challenging tool-use tasks across two different hands, UniDex-VLA achieves 81\% average task progress and outperforms prior VLA baselines by a large margin, while exhibiting strong spatial, object, and zero-shot cross-hand generalization. Together, UniDex-Dataset, UniDex-VLA, and UniDex-Cap provide a scalable foundation suite for universal dexterous manipulation.

\end{abstract}

\section{Introduction}
\label{sec:intro}

In recent years, learning from demonstrations~\cite{dp,dp3,zhao2023learning,xue2025reactive,kim2024openvla,black2024pi_0,bjorck2025gr00t} has become the de facto paradigm for visuomotor control, enabling robots to acquire complex skills and motion patterns. However, achieving general, human-level manipulation under supervised learning remains challenging. Collecting real-robot demonstrations is labor-intensive and scales poorly, creating a persistent data bottleneck. Moreover, most robot foundation policies focus on parallel-jaw grippers, while foundation models for dexterous hands remain scarce—even though everyday tool-use often requires dexterous hands and many tasks (e.g., using scissors or spray bottles) are infeasible with grippers.

Simply porting gripper-based VLA designs to dexterous hands is insufficient. Building foundation models for dexterous hands is substantially more challenging than for grippers. The key difficulties are: (i) dexterous hand data are harder to collect than gripper data, and large, broadly usable pretraining datasets remain limited; (ii) dexterous hands vary widely in DoFs, morphology, kinematics, and appearance, leading to poor transfer of data and policies across hands; and (iii) dexterous hand control is inherently high-dimensional, demanding expressive action spaces and effective learning algorithms.

To address pretraining data scarcity, we leverage the fact that dexterous robot hands are designed to mimic human hands and often share similar action patterns, while humans naturally generate abundant manipulation data in daily life. Egocentric human videos are cheaper, more diverse than robot teleoperation data and easier to scale. We therefore transform human videos into robot-executable trajectories to build a robot-centric dataset from human activity. However, there are substantial \emph{kinematic} and \emph{visual} gaps between human and robot hands. To close these gaps, we (i) introduce a human-in-the-loop retargeting procedure that combines fingertip-based inverse kinematics with interactive adjustment to align robot fingertip trajectories with human trajectories, ensuring physically plausible hand–object contacts; and (ii) mask the human hand in the visual stream and attach the retargeted robot hand into scene pointclouds to reduce visual mismatch.

Following this human-to-robot transformation pipeline, we construct \textbf{UniDex-Dataset} by building on open-source egocentric RGB-D manipulation videos~\cite{kwon2021h2o,banerjee2024hot3d,liu2024taco,liu2022hoi4d}. \textbf{UniDex-Dataset} is a unified foundation dataset comprising 9M paired image–pointcloud–action frames and over 50K trajectories across eight dexterous hand platforms, covering active DoFs from 6 to 24. To our knowledge, UniDex-Dataset is the first dataset to span such a broad spectrum of dexterous hand morphologies at this scale. We also provide protocols that allow researchers to contribute new hands or human datasets with minimal effort, continually scaling UniDex-Dataset and accelerating progress on dexterous manipulation.

To tackle heterogeneous embodiments and high-dimensional control, we further define a unified action space, the \textbf{Function–Actuator–Aligned Space (FAAS)}, which maps functionally similar actuators to shared coordinates. FAAS provides a function-centric control interface and enables skill transfer across different hands. Building on FAAS, we train \textbf{UniDex-VLA}, a 3D vision–language–action policy pretrained on UniDex-Dataset and finetuned with task demonstrations, serving as a foundation model that supports diverse dexterous hands.

In addition, we design a portable human-data capture setup, \textbf{UniDex-Cap}, which records synchronized RGB-D streams and human hand poses and converts them into robot-centric trajectories via the same transformation pipeline. UniDex-Cap enables efficient co-training on transformed human data together with smaller amounts of robot data, reducing teleoperation cost while preserving performance.

We evaluate UniDex-VLA on five challenging real-world tool-use tasks across two different hands. Across these tasks, \textbf{UniDex-VLA} achieves strong performance, outperforming other VLA baselines by a large margin (e.g., 81\% average task progress vs.\ $\pi_0$~\cite{black2024pi_0} at 38\%), and demonstrates strong spatial, object, and cross-hand generalization; with FAAS and pretraining, it transfers skills to unseen hands in a zero-shot manner. Leveraging UniDex-Cap, we also provide a quantitative study showing how transformed human data can reduce post-training costs via human–robot co-training.

Our contributions are summarized as follows:
\begin{itemize}
  \item \textbf{UniDex-Dataset:} a unified, diverse dexterous hand dataset (9M paired frames, over 50K trajectories, 8 hands, 6–24 DoFs) that supports large-scale pretraining toward universal dexterous hand foundation models.
  \item \textbf{FAAS \& UniDex-VLA:} a function–actuator–aligned unified action space and a pretrained 3D vision–language–action model that achieves state-of-the-art performance on real-robot benchmarks, with strong spatial, object, and cross-hand generalization.
  \item \textbf{Human–Robot Data Co-training with UniDex-Cap:} a simple portable capture setup and pipeline that support human–robot data co-training; we quantitatively study how transformed human data can partially substitute real-robot demonstrations during post-training, showing that egocentric human videos both scale pretraining and reduce real-robot data needs.
\end{itemize}

\section{Related Work}
\label{sec:related-work}

\subsection{Dexterous Manipulation}
Early research on dexterous manipulation was grounded in analytic and classical control formulations~\cite{ponce1997computing,mordatch2012contact,bai2014dexterous,kerr1986analysis,arimoto2004intelligent}, and has since progressed toward learning-based methods that enable in-hand reorientation, rotation, and grasping~\cite{qi2023general,yin2023rotating,lin2025learning,chen2023visual,akkaya2019solving,fang2025anydexgrasp,zhong2025dexgraspvla,he2025dexvlg,yuan2024learning,zhong2025dexgrasp,si2024difftactile,jian2025g,wang2025dexh2r,ding2024preafford,  zhang2023flexible}. Despite these advances, most approaches are tailored to specific tasks (grasping) or hardware and struggle to generalize to everyday tool-use. In contrast, we present UniDex-VLA, a foundation model aimed at general-purpose dexterous hand control.

\subsection{Robot Foundation Policies and Unified Action Space}

Diffusion-based policies and their variants constitute strong imitation-learning baselines~\cite{dp,dp3,xue2025reactive,tian2025pdfactor,xu2025diffusion}. With the rise of LLMs and VLMs, vision–language–action (VLA) models~\cite{zhang2025elucidating,kim2024openvla,black2024pi_0,bjorck2025gr00t,miao2025fedvla,ji2025robobrain,liu2024rdt,bu2025univla,zhang2025align,zhang2025robochemist} further scale imitation learning, but most existing approaches are pretrained on large-scale gripper-centric datasets. Recent efforts toward dexterous VLAs~\cite{zhong2025dexgraspvla,he2025dexvlg} leverage simulation or limited real-world data, typically focusing on grasping and relying on hand-specific representations. In contrast, UniDex-VLA is pretrained on UniDex-Dataset to serve as a unified foundation policy for more general dexterous manipulation.

Designing a unified action space for robot foundation policies to handle embodiment heterogeneity is crucial for cross-embodiment generalization. RDT-1B~\cite{liu2024rdt} preserves the semantic structure of control signals, while $\pi_0$~\cite{black2024pi_0} adopts a left-aligned action representation, and other methods introduce latent action spaces~\cite{zhang2025align,bu2025univla}. However, these approaches primarily target gripper-centric actions. EgoVLA~\cite{yang2025egovla} attempts to leverage human parameters as a dexterous representation, but requires inverse kinematics in the post-training stage, which introduces additional errors, particularly for high-DoF dexterous hands. In contrast, FAAS provides a function-centric unified action representation that is post-processing-free, enabling more reliable cross-hand skill transfer.

\subsection{Learning from Human Videos}
Learning from human videos mitigates the data cost bottleneck but introduces visual and kinematic domain gaps. Prior work uses human hand trajectories for planning or control~\cite{qin2022dexmv,wang2023mimicplay,yuan2024general,wen2023any,liu2025egozero,li2025learning,chen2025vidbot}; others apply retargeting with sim-to-real pipelines~\cite{yuan2025hermes,li2025maniptrans,chen2024object} or human-in-the-loop corrections~\cite{wang2024dexcap}, and some co-train with robot data~\cite{kareer2025egomimic,yuan2025motiontrans,qiu2025humanoid,zhou2025mitigating} to bridge the gap. However, many such pipelines primarily target grippers or do not scale robustly. There are also approaches that pretrained on egocentric human videos without explicit supervision of hand motion~\cite{nair2022r3m,ye2024latent,zeng2024learning,niu2025pre}. More recent methods pretrain foundation models on egocentric videos to predict human hand motion, followed by specialized post-training to align with robot actions~\cite{yang2025egovla,luo2025being}, however these additional alignment stages can be complex and brittle. Our approach instead generates robot-centric dexterous hand supervision for pretraining, removing the need for specialized alignment tricks during fine-tuning while maintaining cross-hand control.

\section{UniDex-Dataset}

\subsection{Overview}
\label{sec:dataset}
\textbf{UniDex-Dataset} is derived from four RGB-D egocentric human-manipulation datasets—H2O~\cite{kwon2021h2o}, HOI4D~\cite{liu2022hoi4d}, HOT3D~\cite{banerjee2024hot3d}, and TACO~\cite{liu2024taco}. We annotate language instructions if needed, segment videos into trajectory clips aligned with those instructions, and filter out invalid segments.
\begin{figure}[h]
  \centering
  \includegraphics[width=0.45\textwidth]{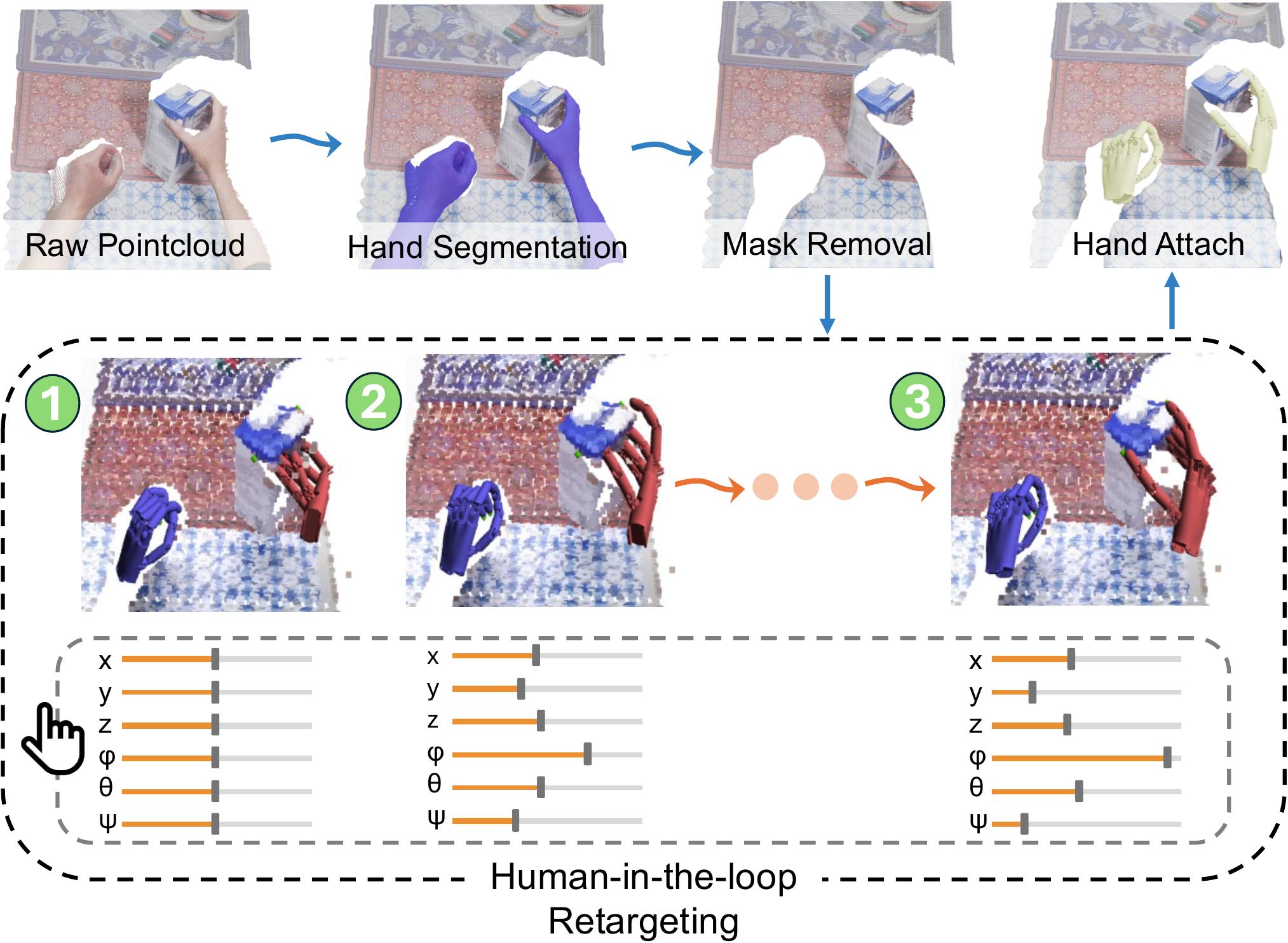}
  \caption{The figure illustrates the complete human–robot transformation pipeline. Starting from the raw scene pointcloud, we first mask out the human hands. We then perform human-in-the-loop retargeting through a user-friendly GUI in which the user only needs to \textbf{adjust slider bars} to modify the dummy base offset. \textcircled{1} shows the retargeted result without adjustment, whereas \textcircled{3} shows the final configuration with improved, more plausible hand–object contact. Finally, after kinematic retargeting, we attach the retargeted robot dexterous hands to the scene.}
  \vspace*{-10pt}
  \label{fig:retarget}
\end{figure}

\begin{figure*}[t]
\centering
  \includegraphics[width=0.95\textwidth]{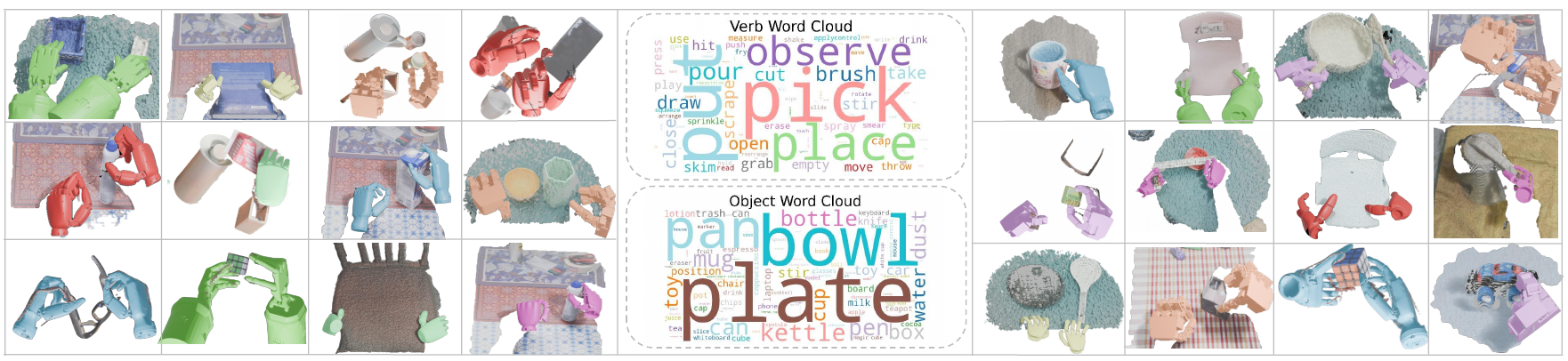}
  \caption{\textbf{UniDex-Dataset visualization.} We show a verb–object word cloud and a subset of UniDex-Dataset. \textbf{Colors} denote \textbf{different hands} (arbitrarily assigned; black corresponds to pretraining data). UniDex-Dataset spans diverse everyday tasks across a wide range of dexterous hand embodiments, including using a mobile phone, opening a milk carton, stir-frying with a spatula, lifting a chair, solving a Rubik’s cube, and more.}
  \label{fig:unidex-dataset}
 \vspace*{-15pt}
\end{figure*}

The transformation from human data to robot-executable trajectories is illustrated in Fig.~\ref{fig:retarget} and detailed in the next subsection. Applying this pipeline, we construct \textbf{UniDex-Dataset} comprising 9M paired image–pointcloud–action frames (recorded at 30~fps) and over $50$k trajectories across eight dexterous hand platforms (Inspire, Leap, Shadow, Allegro, Ability, Oymotion, Xhand, and Wuji), covering active DoF from 6 to 24. Figure~\ref{fig:unidex-dataset} visualizes the verb–object word cloud for the dataset and a subset of the data, spanning diverse daily manipulation tasks such as using a mobile phone, opening a milk carton, and stir-frying with a spatula. Table~\ref{tab:dex-datasets-wide} compares UniDex-Dataset with released collected dexterous manipulation datasets~\cite{fourier2025actionnet,liu2024realdex,wu2024robomind} along the axes of trajectory count, hand variety and scene diversity, and supported perception modalities, highlighting the advantages of UniDex-Dataset. Owing to its diversity and robot-centric formulation—i.e., with minimal embodiment gap to the post-training stage—UniDex-Dataset serves as a strong foundation for pretraining dexterous manipulation models.

\subsection{Human-Robot Transformation}
\label{sec:transform}
Transforming human data into robot trajectories requires overcoming two core gaps: \emph{kinematic} and \emph{visual}. We outline our methods below.

\subsubsection{Kinematic Retargeting}
Fingertips are the primary contact points in human–object interaction. Our goal is to align human fingertip trajectories with those of the robot hand in 3D, while allowing a global hand-base adjustment to better ensure physically plausible contact.

Given a human hand pose, we extract $m$ fingertip targets
\begingroup
\setlength{\abovedisplayskip}{6pt}
\setlength{\belowdisplayskip}{6pt}
\begin{equation}
X^\star = \big[x_1^\star,\ldots,x_m^\star\big] \in \mathbb{R}^{3 \times m},
\end{equation}
\endgroup
where $m$ equals the number of robot fingers. The global human hand transform in the world frame is $T_{\text{hand}}$.

To precisely apply fingertip-based IK while permitting a base adjustment, we introduce a \emph{6-DoF alignment offset}, implemented as a \textbf{dummy base} inserted before the real robot base. Let $T_{\text{offset}}$ be the rigid transform from the dummy base to the real base, and let $T_{\text{world}}^{\text{dummy}}$ be the dummy-base pose in the world frame. The forward kinematics of fingertip $i$ is
\begingroup
\setlength{\abovedisplayskip}{6pt}
\setlength{\belowdisplayskip}{6pt}
\begin{equation}
x_i(q; T_{\text{offset}}) \;=\;
\operatorname{Trans}\!\left(
T_{\text{world}}^{\text{dummy}}\,
T_{\text{offset}}\,T_i(q)
\right) \in \mathbb{R}^3,
\end{equation}
\endgroup
where $T_i(q)$ is the homogeneous transform from the robot base to fingertip $i$, and $\operatorname{Trans}(\cdot)$ extracts the translation. We set $T_{\text{world}}^{\text{dummy}} = T_{\text{hand}}$ and keep it fixed during optimization. Stacking fingertip residuals yields the IK error:
\begingroup
\setlength{\abovedisplayskip}{6pt}
\setlength{\belowdisplayskip}{6pt}
\begin{equation}
e(q, T_{\text{offset}}) =
\begin{bmatrix}
x_1(q; T_{\text{offset}}) - x_1^\star\\
\vdots\\
x_m(q; T_{\text{offset}}) - x_m^\star
\end{bmatrix}
\in \mathbb{R}^{3m}.
\label{eq:retarget}
\end{equation}
\endgroup

%(\texttt{calculateInverseKinematics2})
% to minimize the following objective:
% \begin{equation}
% \label{eq:objective}
% \min_{q}~
% \frac{1}{2}\,\|e(q, T_{\text{offset}})\|_2^2
% + \frac{1}{2}\,\|\Gamma^{1/2}(q - q_{\text{rest}})\|_2^2
% \quad
% \text{s.t. }~ \ell \le q \le u,
% \end{equation}
% where $q_{\text{rest}}$ (\texttt{restPoses}) is the nominal ``open-hand'' configuration
% used as a soft regularization target.
% The joint limits $\ell, u$ correspond to \texttt{lowerLimits} and \texttt{upperLimits} in the URDF.

% The diagonal regularization matrix
% $\Gamma = \mathrm{diag}(\gamma_1, \ldots, \gamma_n) \succeq 0$
% is derived directly from the joint damping and range parameters:
% \begin{equation}
% r_j = u_j - \ell_j, \quad
% d_j = \texttt{jointDamping[j]}, \quad
% \gamma_j = \alpha_d \frac{d_j}{r_j^2 + \varepsilon},
% \label{eq:gamma}
% \end{equation}
% where $\alpha_d > 0$ controls overall regularization strength and $\varepsilon > 0$
% prevents numerical instability when $r_j$ is small.
% This scaling normalizes the damping influence by joint range:
% larger $r_j$ weakens regularization, while larger $d_j$ strengthens it.

\begin{table*}[h]
\centering
\small
\setlength{\tabcolsep}{6pt}
\renewcommand{\arraystretch}{1.15}
\begin{tabular}{lcccccccc}
\hline
Dataset &
\shortstack[c]{\# of\\Trajectories} &
\shortstack[c]{\# of\\Hands} &
% \shortstack[c]{\# of\\Verbs} &
\shortstack[c]{Language\\Annotations} &
\shortstack[c]{Varied\\Scenes} &
RGB & Depth & {Pointcloud} \\
\hline
UniDex-Dataset & 52K & 8 & \cmark & \cmark & \cmark & \cmark & \cmark \\
ActionNet~\cite{fourier2025actionnet} & 30K & 2 & \cmark & \xmark & \cmark & \cmark & \xovercmark \\
RoboMind~\cite{wu2024robomind} & 19K & 1 & \cmark & \xmark & \cmark & \cmark & \xmark \\
RealDex~\cite{liu2024realdex} & 2K & 2 & \cmark & \xmark & \cmark & \cmark & \cmark \\
\hline
\end{tabular}
\caption{\textbf{Comparison between UniDex-Dataset and other dexterous manipulation datasets.} UniDex-Dataset advances in total trajectories, variety across hands/actions/scenes, and supports for all perception modalities. \protect\xovercmark denotes the pointcloud in ActionNet~\cite{fourier2025actionnet} is very low-quality.}
\vspace*{-10pt}
\label{tab:dex-datasets-wide}
\end{table*}

For robot hands containing \emph{mimic joint structures} (e.g., Inspire, Oymotion, Agility),
we handle dependent joints through an iterative correction process.
After solving the primary IK problem, each mimic joint $j_s$ is updated from its master joint $j_m$ as
\begingroup
\setlength{\abovedisplayskip}{6pt}
\setlength{\belowdisplayskip}{6pt}
\begin{equation}
q_{j_s} = k\, q_{j_m} + c
\end{equation}
\endgroup

consistent with the kinematic model specification, where $k$ and $c$ denote the mimic constraints. This correction is repeated for $N$ iterations, re-evaluating fingertip error each time until convergence.

% We adopt a \textbf{two-stage, human-in-the-loop retargeting procedure:}

For implementation, we provide a user-friendly and rapid process. The whole pipeline is a two-stage, human-in-the-loop retargeting procedure.

\begin{enumerate}[leftmargin=1.2em]
    \item \textbf{Automatic stage.}~
    Given an initial $T_{\text{offset}}$, we solve Eq.~\ref{eq:retarget} via PyBullet~\cite{coumans2016pybullet}’s
    multi-end-effector IK solver to obtain a joint configuration $q$
    that minimizes fingertip error while satisfying joint limits and damping.

    \item \textbf{Interactive stage.} A lightweight GUI exposes the six degrees of freedom of $T_{\text{offset}}$
    (three translations and three rotations, as shown in Fig.~\ref{fig:retarget}) and other configuration for IK solver.
    The user visually inspects alignment and manually adjusts $T_{\text{offset}}$;
    after each adjustment, we re-solve the IK problem.
    This process typically converges within a few manual tweaks, producing robust
    fingertip alignment across diverse poses. \textcircled{1} and \textcircled{3} in Fig.~\ref{fig:retarget} shows the comparison between and after the interactive stage.
    
\end{enumerate}
% Crucially, the human-in-the-loop retargeting is calibrated once per (dataset, hand) pair rather than per trajectory. For each human dataset and target robot hand, we perform a single interactive session in our GUI to select dummy-base offsets and a small set of retargeting hyperparameters, and then apply this configuration to all trajectories for that hand, with only occasional spot checks. This is feasible because each human dataset is typically collected with a fixed capture setup and hand–pose estimation pipeline, yielding relatively consistent wrist poses and fingertip trajectories across videos. As a result, most human–robot discrepancies appear as systematic biases (e.g., global shifts or rotations) that can be corrected by one-time calibration, while residual variations remain small. In practice, this procedure has been sufficient to cover most of a dataset for each hand without further manual adjustment, enabling our transformation pipeline to scale to large egocentric datasets with modest human effort.
For each human dataset and each dexterous hand, we perform a basic interactive calibration to select dummy base offsets to handle systematic differences across datasets (e.g., coordinate frames/ hand-pose estimation bias) and hand morphology differences. We then adjust a small subset of frames, focusing on contact-rich segments to improve contact plausibility. In practice, we find the basic calibration suffices to cover the vast majority of trajectories, enabling our transformation pipeline to scale to large egocentric datasets with modest human effort.

\subsubsection{Visual Alignment}
We compute pointclouds from RGB-D frames. Then to reduce the visual gap, we mask human hands (using WiLoR~\cite{potamias2025wilor} together with SAM2~\cite{ravi2024sam}) and remove the corresponding points. We then place the retargeted robot-hand mesh into the scene and render its geometry into the pointcloud. Finally, we reproject the fused pointcloud back to the RGB-D frame via a pinhole camera model~\cite{hartley2003multiple} to avoid occlusions caused by incorrect depth ordering, matching the single-view setting used during real-world fine-tuning.

% \section{Final copy}

% You must include your signed IEEE copyright release form when you submit your finished paper.
% We MUST have this form before your paper can be published in the proceedings.

% Please direct any questions to the production editor in charge of these proceedings at the IEEE Computer Society Press:
% \url{https://www.computer.org/about/contact}.

\section{UniDex-VLA}
\label{sec:unidexvla}

\subsection{Unified Action Space: FAAS}
We pretrain our robot foundation model on UniDex-Dataset, which spans diverse dexterous-hand embodiments. A unified action space that enables transfer across hands is therefore critical. To this end, we introduce a simple yet effective action representation, the \textbf{Function–Actuator–Aligned Space} (\textbf{FAAS}). For any dexterous hand with $n$ actuated DoFs in its kinematic model, each \emph{actuator} is mapped to the FAAS \emph{index} corresponding to its functional role. Here we use "actuator" broadly to denote any controllable DoF/channel derived from the robot URDF, including mimic joints when present.

\begin{figure}[h]
\centering
  \includegraphics[width=0.45\textwidth]{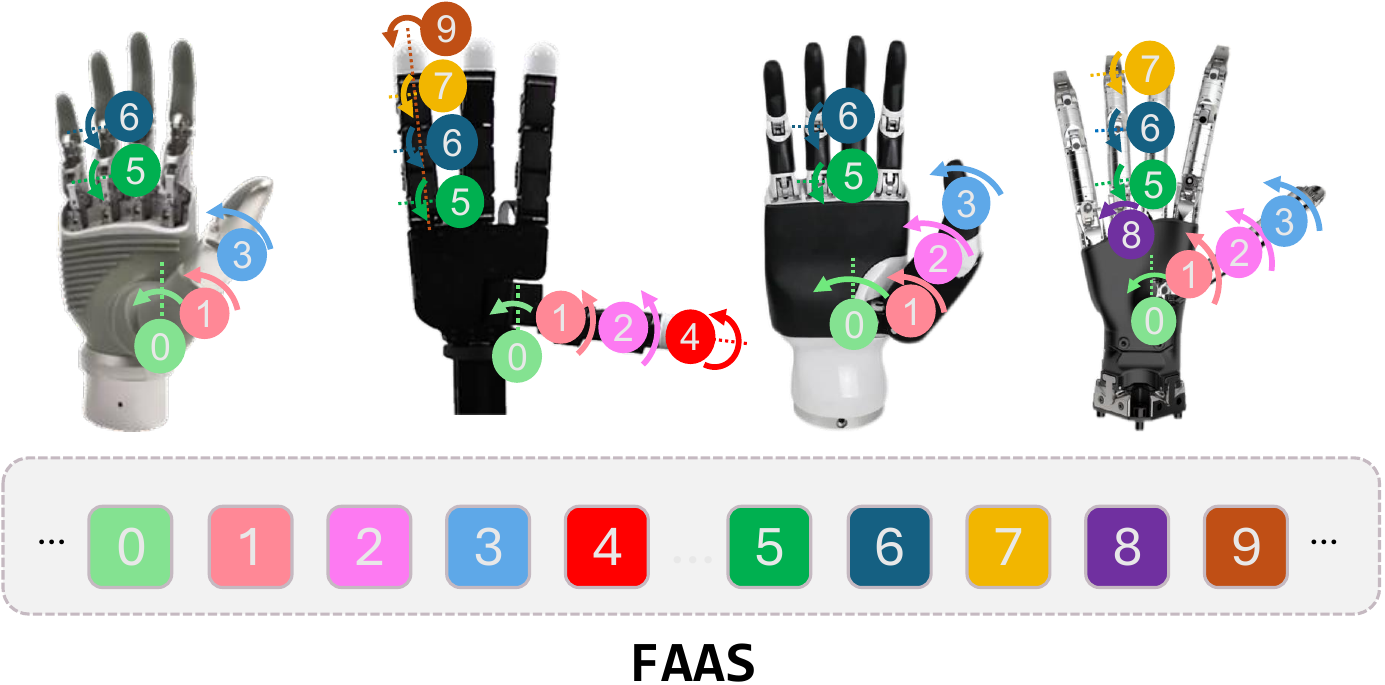}
\caption{\textbf{Function–Actuator–Aligned Space (FAAS).} We show the thumb and ring fingers of Oymotion (11 actuators), Allegro (16), Inspire (12), and Wuji (20), with colors denoting individual joints, curves indicating rotation directions, and dotted lines indicating rotation axes. Indices \{0,1,3,5,6\} are aligned across all four hands because the corresponding joints share similar functional roles.}
\label{fig:faas}
\vspace*{-15pt}
\end{figure}

Conceptually, FAAS exposes a function-centric control interface shared across embodiments rather than a URDF-specific joint space. Although dexterous hands differ in link lengths, couplings, and layouts, they all implement a small set of functional primitives—such as thumb–index pinch, finger curling around handles, or lateral ab-/adduction for stabilization. FAAS groups actuators by these functional roles and maps them into a common coordinate system, discarding embodiment-specific nuisance factors while preserving task-relevant control semantics. Fig.~\ref{fig:faas} illustrates, for the thumb and ring fingers of different hands, how individual joints are mapped to FAAS indices. 

FAAS is an 82-dimensional action vector. The first 18 dimensions encode wrist poses (9 per hand), where each 9d pose consists of a 6d continuous rotation representation (two 3d vectors for the local $x$- and $y$-axes) followed by a 3d translation. and the remaining 64 dimensions encode joint commands, with 32 slots for each hand. Among these slots, we reserve 21 \emph{base} actuator slots that are shared across all hands, and use the remaining slots for hand-specific DoFs (e.g., additional wrist joints on the Shadow Hand) and for future hands. The details of joint mapping for different hands are shown in Sec.~\ref{sec:ap-faas} and Fig.~\ref{fig:detailed_faas} in Appendix.

% FAAS further encodes wrist poses in the first 18 dimensions (9 per hand), and allocates 32 slots for each hand’s finger actuators, yielding an 82-dimensional action vector.

% FAAS is an 82-dimensional action vector. The first 18 dimensions encode wrist poses (9 per hand), and the remaining 64 dimensions encode joint commands, with 32 slots for each hand. Among these slots, we reserve 21 \emph{base} actuator slots that are shared across all hands, and use the remaining slots for hand-specific DoFs (e.g., additional wrist joints on the Shadow Hand) and for future hands. By mapping functionally similar actuators to the same FAAS coordinates, policy knowledge transfers more readily across embodiments.

% where each 9d pose consists of a 6d continuous rotation representation (two 3d vectors for the local $x$- and $y$-axes) followed by a 3d translation. 

\begin{figure*}[t]
  \centering
  \includegraphics[width=0.98\textwidth]{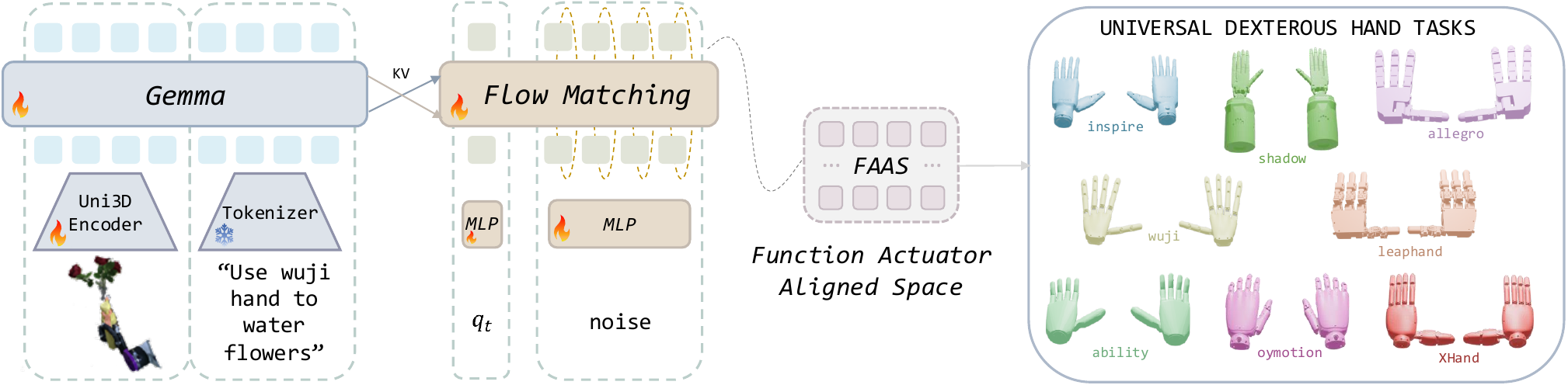}
  \caption{\textbf{Overview of UniDex-VLA.} At time $t$, the model consumes a single-view colored pointcloud $P_t$, a language instruction $\ell_t$, and proprioception $q_t$, and predicts an $H$-step action chunk $A_t = [a_t,\ldots,a_{t+H-1}]$ expressed in the unified action space FAAS. Uni3D~\cite{zhou2023uni3d} encodes the colored pointcloud; features are fused with text and proprioception in the backbone and decoded into FAAS actions. The policy is pretrained on UniDex-Dataset and optimized with a conditional flow-matching objective.}
  \label{fig:framework}
  \vspace*{-15pt}
\end{figure*}

\subsection{VLA Policy}
UniDex-VLA aims to be a 3D, language-conditioned foundation model for dexterous control. Unlike prior VLAs that pair 2D encoders with low-dimensional gripper actions, our setting is inherently volumetric and high-DoF: tool-use requires reasoning about fine 3D geometry and contact affordances, especially in the egocentric single-view observation. By coupling 3D visual inputs with the unified FAAS action space, UniDex-VLA aligns geometric perception and control in a shared representation, supporting spatial, object, and cross-hand generalization.

\subsubsection{Observations and Action Outputs}
As shown in Fig.~\ref{fig:framework}, the observation at time $t$ is
$o_t = [P_t,\, \ell_t,\, q_t]$, where $P_t$ is a single-view colored pointcloud derived from an RGB-D image and then cropped and downsampled, $\ell_t$ is a natural-language instruction, and $q_t$ is a vector of robot proprioceptive states. We model $p(A_t \mid o_t)$, where $A_t = [a_t, \ldots, a_{t+H-1}]$ denotes an $H$-step action chunk~\cite{zhao2023learning}. Both $q_t$ and each $a_t$ are represented in FAAS. For the wrist in $q_t$, we use an \emph{absolute} pose; for action outputs, we adopt a \emph{relative} wrist pose with respect to the first frame of the action chunk, following UMI~\cite{chi2024universal}. For dexterous-hand joints, we likewise use abstracted representations in both $q_t$ and $a_t$.

\subsubsection{Model Architecture}
The UniDex-VLA architecture largely follows $\pi_0$~\cite{black2024pi_0}, with modifications for pointcloud inputs. Specifically, we replace the SigLIP~\cite{zhai2023sigmoid} 2D vision encoder in PaliGemma~\cite{beyer2024paligemma} with Uni3D~\cite{zhou2023uni3d}, a strong 3D pointcloud encoder. Uni3D adopts a vanilla ViT~\cite{dosovitskiy2020image} design and is initialized from a 2D pretrained ViT, aligning pointcloud features with image–text–aligned features. We train the policy with a conditional flow-matching objective and generate denoised action chunks at inference time via forward–Euler integration~\cite{lipman2022flow}.

\noindent More details of UniDex-VLA training are shown in Sec.~\ref{sec:ap-training-detail} in Appendix.

% \subsubsection{Flow-Matching–Based Action Generation}
% We minimize a conditional flow-matching objective for training:
% \begin{equation}
% L^{\tau}(\theta)
% = \mathbb{E}_{p(A_t \mid o_t),\, q(A_t^{\tau} \mid A_t)}\!\left[
% \left\lVert v_{\theta}(A_t^{\tau}, o_t) - u(A_t^{\tau} \mid A_t) \right\rVert
% \right],
% \end{equation}
% where $\tau \in [0,1]$ and $q(A_t^{\tau} \mid A_t)=\mathcal{N}\!\bigl(\tau A_t,\; (1-\tau)I\bigr)$ defines a linear–Gaussian probability path. We sample
% $A_t^{\tau} = \tau A_t + (1-\tau)\epsilon$ with $\epsilon \sim \mathcal{N}(0,I)$ and compute the target conditional vector field
% $u(A_t^{\tau} \mid A_t) = \epsilon - A_t$. The network is trained so that the predicted vector field $v_{\theta}(A_t^{\tau}, o_t)$ approximates $u(A_t^{\tau} \mid A_t)$.

% At inference, we integrate the learned vector field with a forward-Euler scheme to generate a denoised action chunk.

\section{Experiments}
\label{sec:experiments}
\subsection{Experimental Setup}
\begin{figure}[h]
  \centering
  \includegraphics[width=0.4\textwidth]{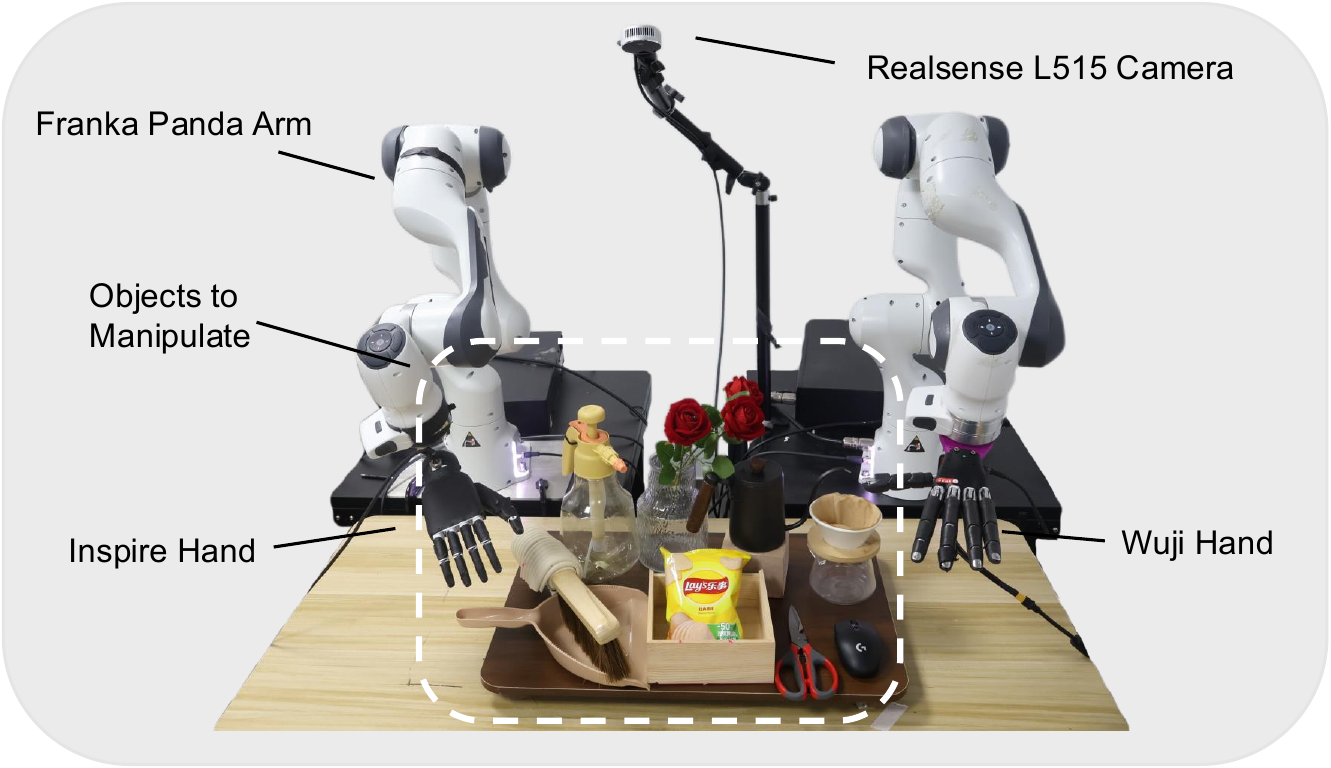}
  \caption{Real-world experiments setup overview}
  \label{fig:real-world setup}
  \vspace*{-10pt}
\end{figure}

 \textbf{Hardware Platform.} Our real-world experiments use a 7-DoF Franka robotic arm equipped with three dexterous end-effectors: an Inspire Hand (6 active, 12 full DoFs), a Wuji Hand (20 active DoFs), and an Oymotion Hand (6 active, 11 full DoFs), all mounted at the end-effector. An Intel RealSense L515 provides egocentric RGB-D observations for all experiments. The complete workstation is shown in Fig.~\ref{fig:real-world setup}.

\begin{figure}[t]
  \centering
  \includegraphics[width=0.47\textwidth]{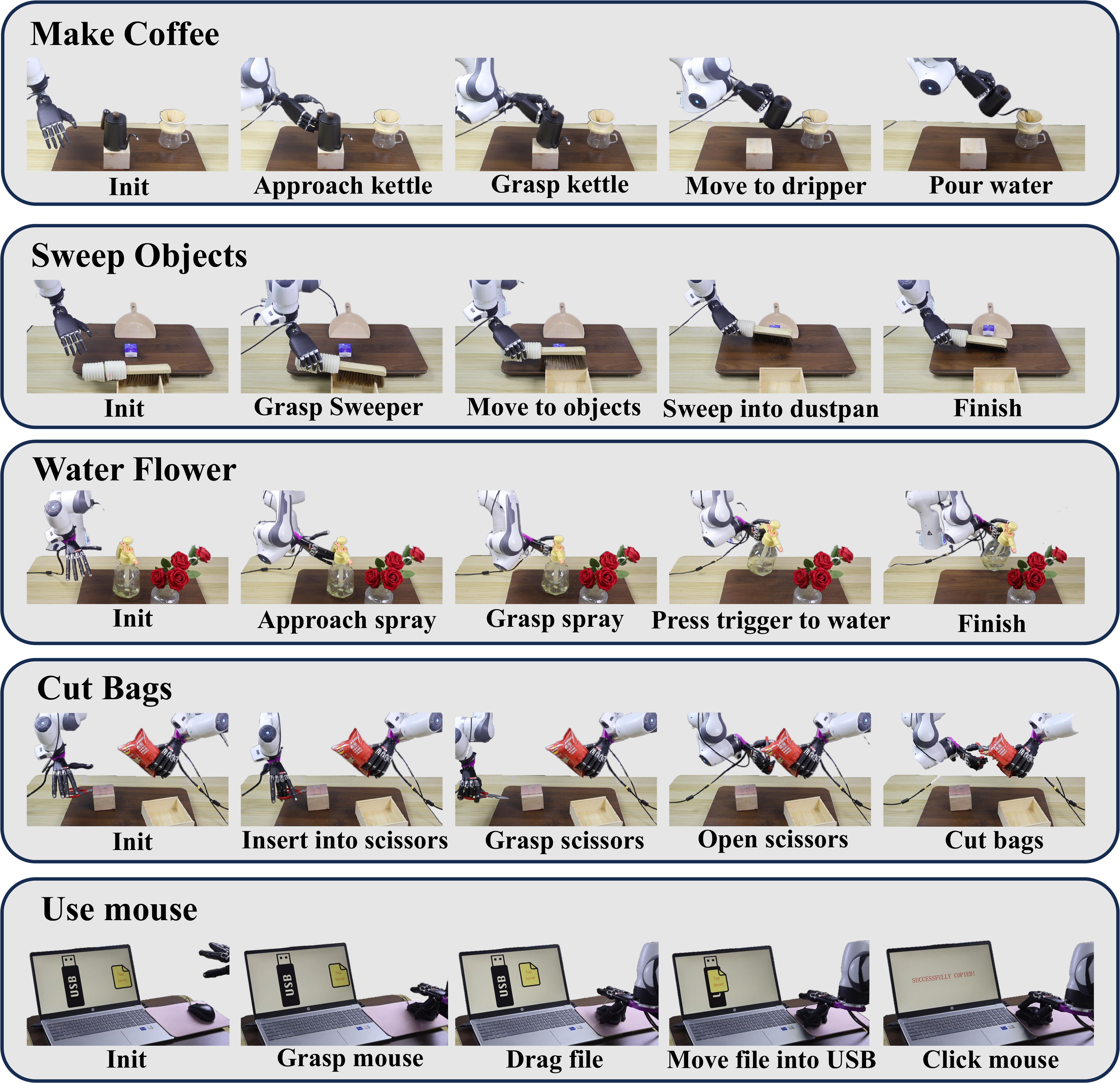}
  \caption{Our real-robot benchmark comprises 5 challenging tool-use tasks. We visualize the key stages of each task, illustrating the precise dexterous control required to successfully complete them.}
  \vspace*{-12pt}
  \label{fig:task-description}

\end{figure}

\noindent\textbf{Task Description.} Everyday manipulation commonly involves many tools designed for human hands—e.g., scissors, spray bottles, and sweepers—which impose stringent requirements on finger coordination and in-hand reconfiguration. To better assess the dexterity and generality of our approach, we evaluate five challenging tool-use tasks, with visualization of different stages in Fig.~\ref{fig:task-description}: (i) \textbf{Make Coffee} (Inspire Hand): Grasp the kettle and lift it to the dripper to pour water to make pour-over coffee. Task decomposed into kettle grasping (\textbf{Grasp}) and water pouring (\textbf{Pour}).
(ii) \textbf{Sweep Objects} (Inspire Hand): Grasp a sweeper and sweep tabletop objects into a dustpan. Task decomposed into sweeper grasping (\textbf{Grasp}) and sweeping (\textbf{Sweep}).
(iii) \textbf{Water Flowers} (Wuji Hand): Grasp a spray bottle, lift it, and press the trigger with the thumb to water flowers. Task decomposed into bottle grasping (\textbf{Grasp}) and pressing trigger to water (\textbf{Press}).
(iv) \textbf{Cut Bags} (Wuji Hand): Insert thumb, middle and ring fingers into scissors and grasp them in a human-like manner to cut bags. Task decomposed into scissors grasping (\textbf{Grasp}) and cutting (\textbf{Cut}).
(v) \textbf{Use Mouse} (Wuji Hand): Place fingers on a computer mouse and use it to drag a file into a USB folder in the desktop interface and click the mouse to finish. We report the mean success rate across all task stages as the \textbf{average task progress}, which serves as our primary metric for comparing methods.

\noindent\textbf{Demonstration Collection.} We build our teleoperation system on OpenTeleVision~\cite{cheng2024open} and dex-retargeting~\cite{qin2023anyteleop} with Apple Vision Pro. We only collect \emph{50} demonstrations per task for fine-tuning.

\noindent\textbf{Baselines.} We compare \textbf{UniDex-VLA} with representative imitation learning and VLA methods: Diffusion Policy (DP)~\cite{dp}, 3D Diffusion Policy (DP3)~\cite{dp3}, and the strong VLA baseline $\pi_0$~\cite{black2024pi_0} pretrained on gripper action datasets. To directly assess the effect of pretraining, we include {UniDex-VLA (No Pretrain)}. We adopt FAAS for UniDex-VLA (No Pretrain) and $\pi_0$, and retain low-dimensional outputs for DP and DP3.

\begin{figure}[t]
  \centering
  \includegraphics[width=0.457\textwidth]{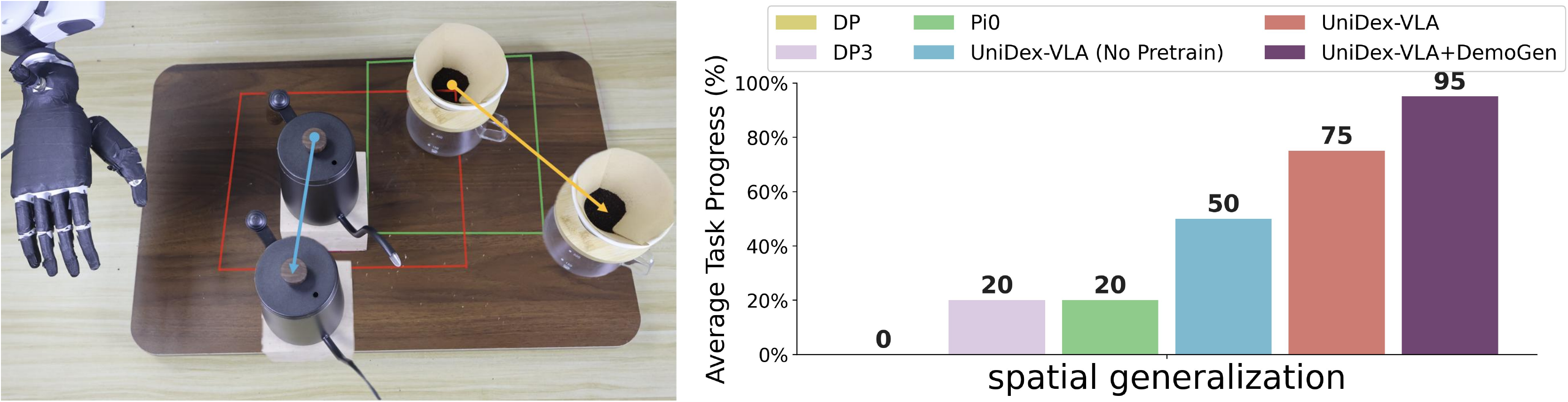}
  \caption{\textbf{Spatial generalization.} Left: the kettle and dripper are placed at \emph{out-of-distribution (OOD)} positions relative to training demonstrations. Red and green lines circling regions denote the training placement ranges for the kettle and dripper, respectively. Right: average task progress for different methods (10 trials each).}
  \label{fig:spatial-generalization}
  \vspace*{-12pt}
\end{figure}

\begin{figure}[t]
  \centering
  \includegraphics[width=0.457\textwidth]{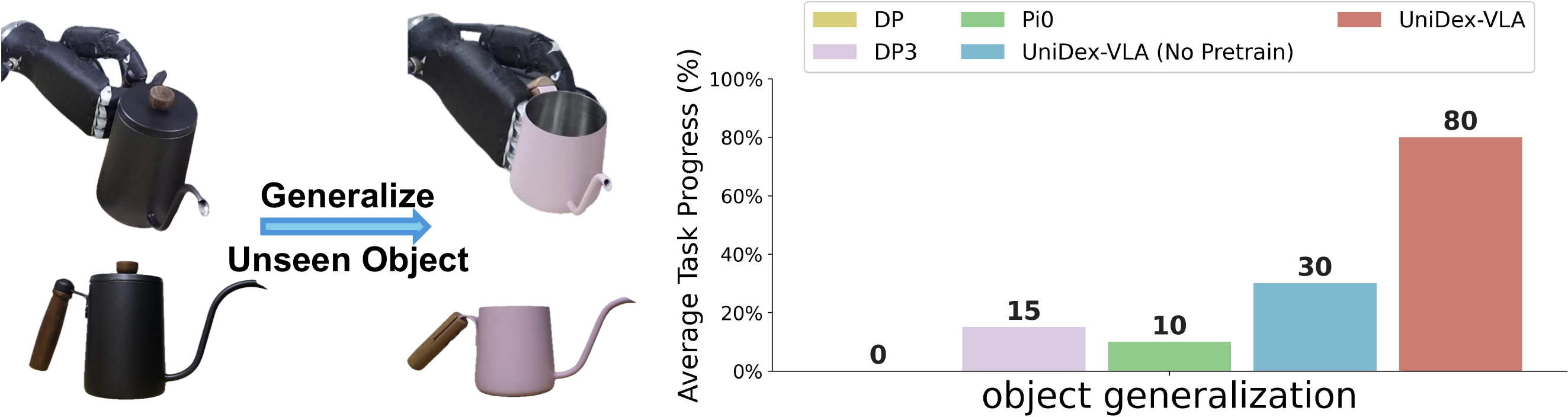}
  \caption{\textbf{Object generalization.} Left: we replace the original black kettle with a smaller purple kettle that differs in color, size, and functional parts (handle \& spout). Right: average task progress for different methods (10 trials each).}
  \vspace*{-12pt}
  \label{fig:obj-generalization}
\end{figure}

\begin{figure}[t]
  \centering
  \includegraphics[width=0.9\linewidth]{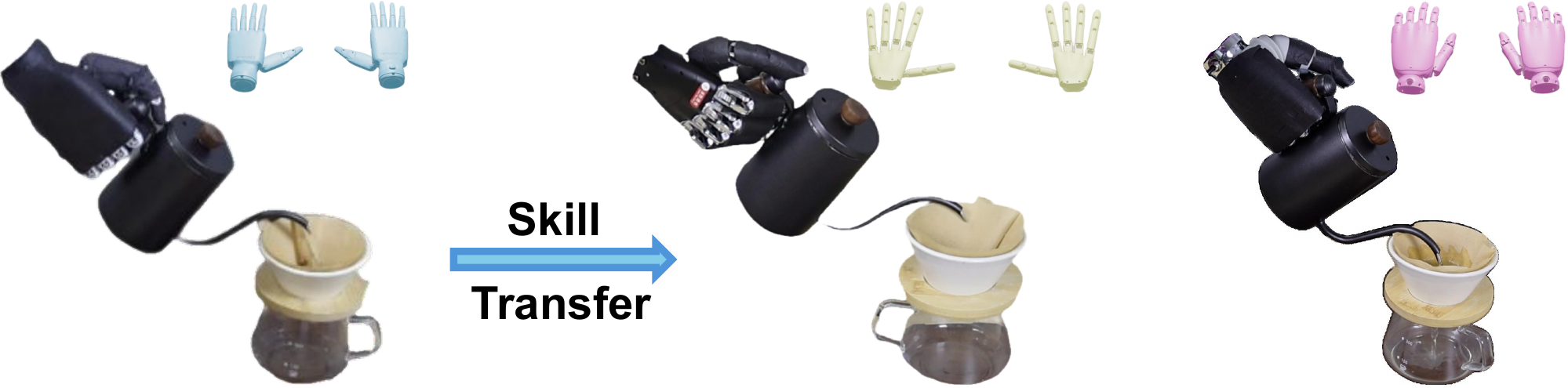}
  % \vspace{4pt}
  {\footnotesize
  \setlength{\tabcolsep}{5pt}
  \renewcommand{\arraystretch}{1.08}
  \begin{tabular*}{0.95\linewidth}{@{\extracolsep{\fill}} l c c c @{}}
    \toprule
    \textbf{Hand Type} & \textbf{$\pi_0$} & \textbf{UniDex-VLA (No Pretrain)} & \textbf{UniDex-VLA} \\
    \midrule
    Wuji     & 0\%  & 0\%  & \textbf{40\%} \\
    Oymotion & 10\% & 5\%  & \textbf{60\%} \\
    \bottomrule
  \end{tabular*}
  }
  \caption{\textbf{Hand generalization (zero-shot skill transfer).} We transfer a policy trained on the Inspire Hand to Wuji and Oymotion. Table reports \textbf{average task progress (\%)} under zero-shot deployment (10 trials each).}
  \vspace*{-15pt}
  \label{fig:hand-generalization}
\end{figure}

\subsection{Performance}
\begin{figure*}[h]
  \centering
  \includegraphics[width=0.90\textwidth]{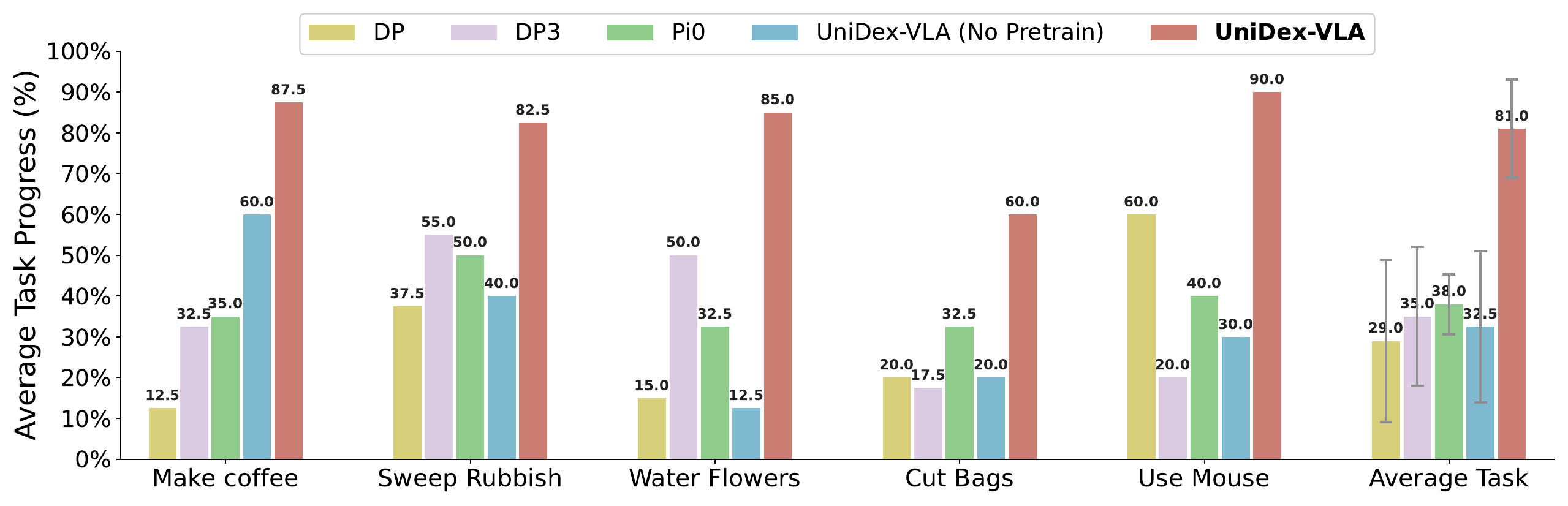}
  \vspace{4pt}
  {\footnotesize
  \setlength{\tabcolsep}{5pt}
  \renewcommand{\arraystretch}{1.08}
  \begin{tabular*}{0.9\linewidth}{@{\extracolsep{\fill}} l c c c c c @{}}
    \toprule
    \textbf{Model} & \textbf{DP} & \textbf{DP3} & \textbf{$\pi_0$} & \textbf{UniDex-VLA (No Pretrain)} & \textbf{UniDex-VLA}\\
    \midrule
    Average Task Progress (avg) & $29.0 \pm 19.9\%$ & $35.0 \pm 17.1\%$ & $38.0 \pm 7.4\%$ & $32.5 \pm 18.5\%$ & $\mathbf{81.0 \pm 12.1\%}$ \\
    Final Success Rate (avg)    & $22.0 \pm 22.5\%$ & $30.0 \pm 18.7\%$ & $35.0 \pm 10.0\%$ & $23.0 \pm 12.0\%$ & $\mathbf{76.0 \pm 17.8\%}$ \\
    \bottomrule
  \end{tabular*}
  }
  \caption{Average task progress across five real-world tasks (top), with aggregate averages of average task progress and final success rate (bottom) over 5 tasks. Each task/algorithm uses \textbf{20} trials.}
  \vspace*{-15pt}
  \label{fig:main-exp-results}
\end{figure*}

We report results on five real-world manipulation tasks across two dexterous hands at Fig.~\ref{fig:main-exp-results}. The results show that, with only 50 demonstrations per task, \textbf{UniDex-VLA} attains high success rates on these challenging, long-horizon tool-use tasks and surpasses all baselines by a large margin, including on the especially difficult \emph{Use Scissors to Cut Bags} task. The performance gap between UniDex-VLA (No-Pretrain) and UniDex-VLA further provides a clear ablation of the benefit of pretraining on UniDex-Dataset. Computing relative improvement over the best competing method (Fig.~\ref{fig:main-exp-results}), UniDex-VLA achieves the largest gain on the hardest setting, \emph{Use Scissors to Cut Bags}, with an \textbf{84.6\%} increase in average task progress. Overall, these results indicate that pretraining endows UniDex-VLA with strong motion priors for dexterous hand control, particularly on highly dexterous tool-use tasks, enabling more efficient adaptation to new and challenging behaviors.

\subsection{Generalization}
Beyond outperforming performance, UniDex-VLA demonstrates strong spatial, object, and hand generalization.

\noindent\textbf{Spatial Generalization.}
UniDex-VLA benefits from 3D perception, and pointclouds further enable simple, automatic data augmentation via geometric editing. In the \textit{Make Coffee} experiment, we segment the pointclouds of the kettle and the dripper, and translate them along the table’s $x$/$y$ axes to sweep across the workspace and generate out-of-distribution (o.o.d.) placements. After editing the pointclouds, the corresponding robot states are aligned to the new scenes using Task and Motion Planning (TAMP)~\cite{dalal2023imitating}. DemoGen~\cite{xue2025demogen} provides an automated pipeline for this procedure. As shown in Fig.~\ref{fig:spatial-generalization}, UniDex-VLA generalizes well across spatial configurations; with DemoGen~\cite{xue2025demogen} augmentation, it approaches very high success rate over full workspace.

\noindent\textbf{Object Generalization.}
As in Fig.~\ref{fig:obj-generalization}, we replace the black kettle with a smaller purple kettle that differs in color, size, and functional parts (handle \& spout). \textbf{UniDex-VLA} maintains strong performance on this unseen object, indicating generalizable tool understanding capacity crucial for robust and general tool-use.

\noindent\textbf{Hand Generalization (Skill Transfer).}
We evaluate cross-hand transfer by taking a policy trained to \emph{Make Coffee} on the Inspire Hand (6 active DoF) and deploying it \emph{zero-shot} on Wuji (20 active DoFs) and Oymotion (6 active DoFs with different kinematics). As shown in Fig.~\ref{fig:hand-generalization}, \textbf{UniDex-VLA} achieves \textbf{60\%} success on Oymotion and \textbf{40\%} on Wuji without any fine-tuning, whereas baselines are near zero. These results highlight that pretraining across diverse dexterous hands—together with FAAS—indeed enables zero-shot cross-hand skill transfer.

\subsection{UniDex-Cap for Human-Robot Data Co-train}
\begin{figure}[t]
  \centering
  \includegraphics[width=0.48\textwidth]{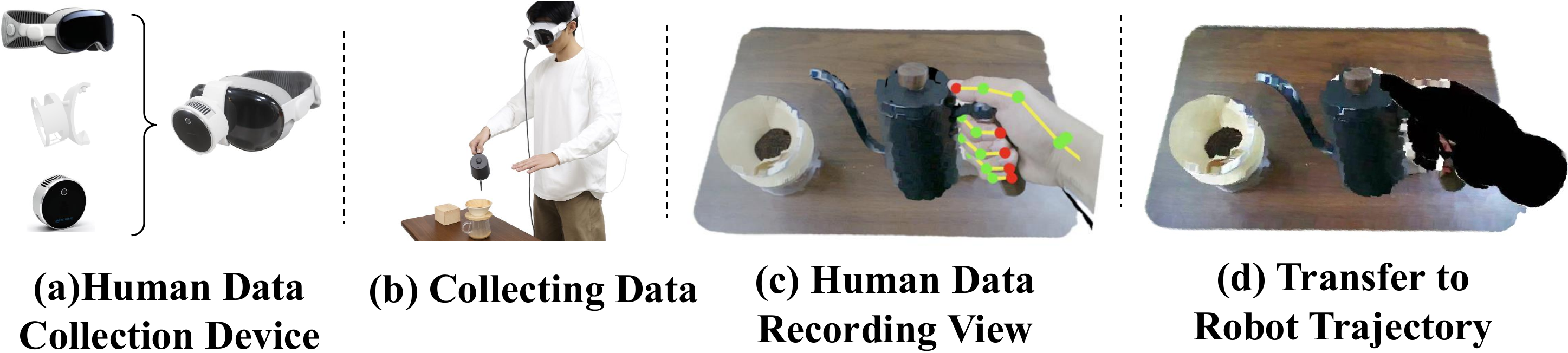}
  \caption{(a,b) show the components of UniDex-Cap. (c,d) shows the example captured data and converted robot-executable trajectories.}
  \label{fig:human_data}
  \vspace*{-15pt}
\end{figure}

We introduce \textbf{UniDex-Cap}, a practical data-capture setup that records synchronized RGB-D streams and hand/head poses. The system combines an Apple Vision Pro for hand and head pose estimation, an Intel RealSense L515 for high-quality RGB-D, and a custom 3D-printed mount to physically couples the two sensors with a fixed rigid transform. This transform is calibrated to ensure the RGB-D stream and the hand/head poses are time-synchronized and expressed in the shared coordinate frame. As illustrated in Fig.~\ref{fig:human_data}, we then apply the human-to-robot transformation pipeline (Sec.~\ref{sec:transform}) to convert captured human data into robot-executable trajectories. In addition, we perform a viewpoint transformation to align human and robot perspectives and downsample the human motion to match typical teleoperation speeds.

\begin{figure}[ht]
  \centering
  \includegraphics[width=0.45\textwidth]{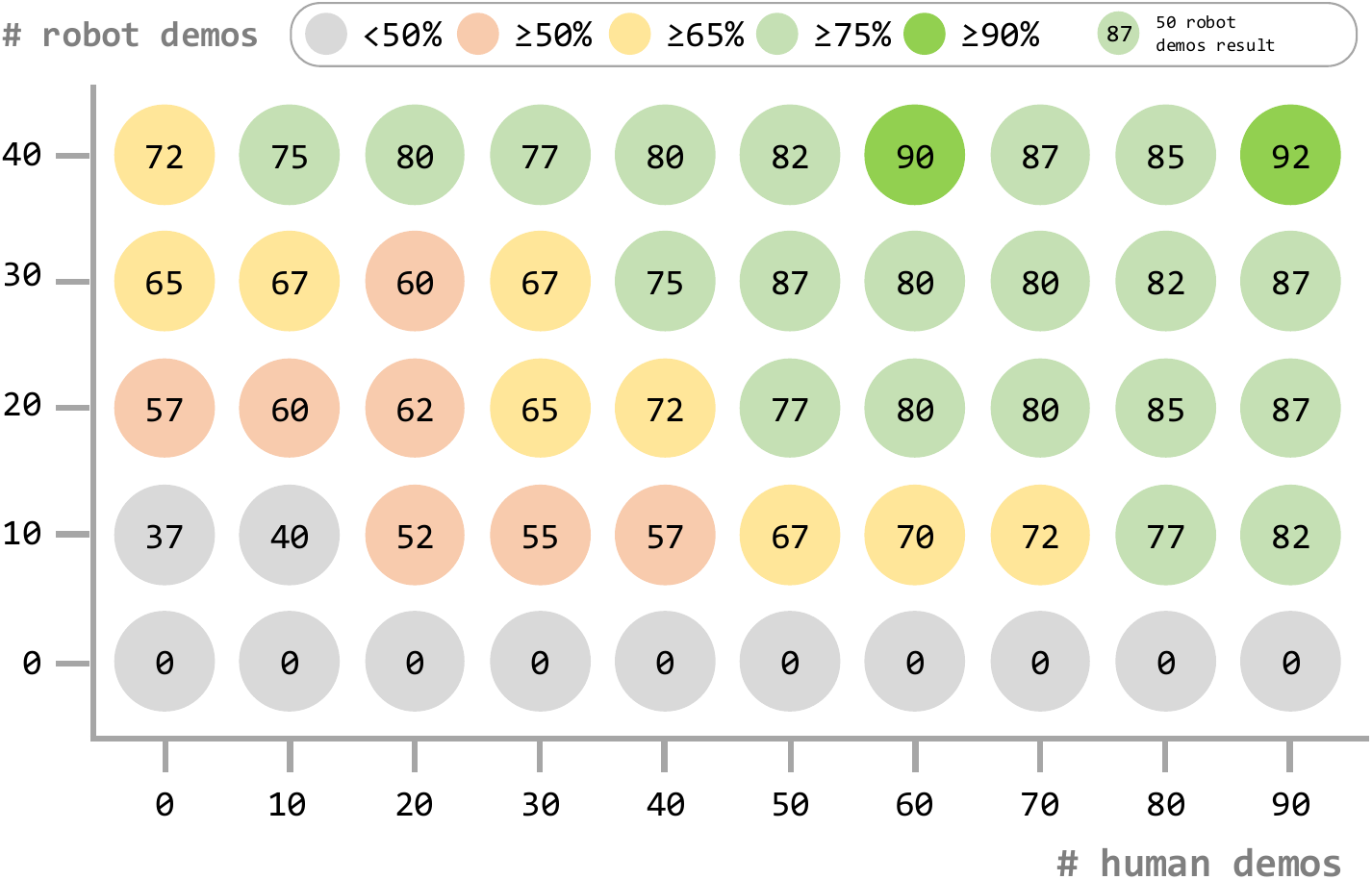}
  \caption{\textbf{Human-Robot co-training.} Average task progress versus the numbers of transformed human demos ($h$) and robot demos ($r$). Colors indicate different performance bands (green: comparable to the $r{=}50$ robot-only result). Each point averages over \textbf{20} trials.}
  \label{fig:human-posttraining}
  \vspace*{-10pt}
\end{figure}

Leveraging UniDex-Cap, we collect human demonstrations, transform them, and \emph{co-train} with real-robot data on \emph{Make Coffee} task to quantitatively explore the effect of human demos during the finetuning stage. Figure~\ref{fig:human-posttraining} reports average task progress versus the numbers of co-trained transformed human demos ($h$) and robot demos. We observe: (i) \textbf{Retargeted human data helps, but robot data is indispensable.} Although for a fixed $r$, increasing $h$ consistently improves average task progress within our evaluated range but success always remains near zero without any robot data. 
(ii) \textbf{Human–robot exchange rate $\approx$ 2:1.} From Fig.~\ref{fig:human-posttraining}, the boundary separating the "high-performance" region (comparable to the $r{=}50$ robot-only result green area) has slope $\approx 2$, suggesting roughly \emph{two human demos can substitute for one robot demo}.
(iii) \textbf{Cost efficiency.} On \emph{Make Coffee} task, human demos are $\sim$5.2$\times$ faster to collect than real robot demos; considering the $\approx$2:1 exchange rate, co-training with human demos can substantially reduce data collection cost.

\section{Conclusion and Limitation}
% We presented \textbf{UniDex}, a robot foundation suite built from egocentric human videos, comprising: \textbf{UniDex-Dataset}, a diverse robot-centric dataset for pretraining universal dexterous hand policies; \textbf{UniDex-VLA}, a FAAS-based 3D VLA policy that achieves strong performance on challenging tool-use tasks with robust spatial, object, and cross-hand generalization; and \textbf{UniDex-Cap}, a portable capture setup that enables cost-efficient human–robot co-training. We hope UniDex can serve as a practical foundation platform for the community, accelerating progress toward general, scalable dexterous manipulation. A key limitation is that our current pipeline does not leverage large \emph{action-free} (or weakly labeled) egocentric activity datasets; extending UniDex to incorporate such data is a promising direction for further scaling dexterous pretraining.
We presented UniDex, a robot foundation suite built from egocentric human videos, comprising UniDex-Dataset, UniDex-VLA, and UniDex-Cap. We believe UniDex can serve as a practical foundation platform for the community, accelerating progress toward general, scalable, and transferable dexterous manipulation. A limitation of our current work is that we do not yet leverage large \emph{action-free} (or weakly labeled) egocentric activity datasets; extending UniDex to incorporate such data is a promising direction for further scaling dexterous pretraining.

\section{Acknowledgment}

We would like to give special thanks to Wuji Technology Inc. for
providing the hardware support, and Hojin Bae, Haoxu Huang, Shaoting Zhu for their technical support. Tsinghua University Dushi Program supports this project.

{
    \small
    \bibliographystyle{ieeenat_fullname}
    \bibliography{main}

@String(CVPR= {IEEE Conf. Comput. Vis. Pattern Recog.})

@String(VR   = {Vis. Res.})

@String(CVPR  = {CVPR})

@article{dp3,
  title={3d diffusion policy: Generalizable visuomotor policy learning via simple 3d representations},
  author={Ze, Yanjie and Zhang, Gu and Zhang, Kangning and Hu, Chenyuan and Wang, Muhan and Xu, Huazhe},
  journal={arXiv preprint arXiv:2403.03954},
  year={2024}
}

@article{dp,
  title={Diffusion policy: Visuomotor policy learning via action diffusion},
  author={Chi, Cheng and Xu, Zhenjia and Feng, Siyuan and Cousineau, Eric and Du, Yilun and Burchfiel, Benjamin and Tedrake, Russ and Song, Shuran},
  journal={The International Journal of Robotics Research},
  pages={02783649241273668},
  year={2023},
  publisher={SAGE Publications Sage UK: London, England}
}

@article{black2024pi_0,
  title={$\pi_0$: A Vision-Language-Action Flow Model for General Robot Control},
  author={Black, Kevin and Brown, Noah and Driess, Danny and Esmail, Adnan and Equi, Michael and Finn, Chelsea and Fusai, Niccolo and Groom, Lachy and Hausman, Karol and Ichter, Brian and others},
  journal={arXiv preprint arXiv:2410.24164},
  year={2024}
}

@article{bjorck2025gr00t,
  title={Gr00t n1: An open foundation model for generalist humanoid robots},
  author={Bjorck, Johan and Casta{\~n}eda, Fernando and Cherniadev, Nikita and Da, Xingye and Ding, Runyu and Fan, Linxi and Fang, Yu and Fox, Dieter and Hu, Fengyuan and Huang, Spencer and others},
  journal={arXiv preprint arXiv:2503.14734},
  year={2025}
}

@article{kim2024openvla,
  title={Openvla: An open-source vision-language-action model},
  author={Kim, Moo Jin and Pertsch, Karl and Karamcheti, Siddharth and Xiao, Ted and Balakrishna, Ashwin and Nair, Suraj and Rafailov, Rafael and Foster, Ethan and Lam, Grace and Sanketi, Pannag and others},
  journal={arXiv preprint arXiv:2406.09246},
  year={2024}
}

@article{chi2024universal,
  title={Universal manipulation interface: In-the-wild robot teaching without in-the-wild robots},
  author={Chi, Cheng and Xu, Zhenjia and Pan, Chuer and Cousineau, Eric and Burchfiel, Benjamin and Feng, Siyuan and Tedrake, Russ and Song, Shuran},
  journal={arXiv preprint arXiv:2402.10329},
  year={2024}
}

@article{xue2025reactive,
  title={Reactive diffusion policy: Slow-fast visual-tactile policy learning for contact-rich manipulation},
  author={Xue, Han and Ren, Jieji and Chen, Wendi and Zhang, Gu and Fang, Yuan and Gu, Guoying and Xu, Huazhe and Lu, Cewu},
  journal={arXiv preprint arXiv:2503.02881},
  year={2025}
}

@article{yang2025egovla,
  title={Egovla: Learning vision-language-action models from egocentric human videos},
  author={Yang, Ruihan and Yu, Qinxi and Wu, Yecheng and Yan, Rui and Li, Borui and Cheng, An-Chieh and Zou, Xueyan and Fang, Yunhao and Cheng, Xuxin and Qiu, Ri-Zhao and others},
  journal={arXiv preprint arXiv:2507.12440},
  year={2025}
}

@article{luo2025being,
  title={Being-H0: Vision-language-action pretraining from large-scale human videos},
  author={Luo, Hao and Feng, Yicheng and Zhang, Wanpeng and Zheng, Sipeng and Wang, Ye and Yuan, Haoqi and Liu, Jiazheng and Xu, Chaoyi and Jin, Qin and Lu, Zongqing},
  journal={arXiv preprint arXiv:2507.15597},
  year={2025}
}

@article{qiu2025humanoid,
  title={Humanoid Policy\~{} Human Policy},
  author={Qiu, Ri-Zhao and Yang, Shiqi and Cheng, Xuxin and Chawla, Chaitanya and Li, Jialong and He, Tairan and Yan, Ge and Yoon, David J and Hoque, Ryan and Paulsen, Lars and others},
  journal={arXiv preprint arXiv:2503.13441},
  year={2025}
}

@article{yuan2025hermes,
  title={Hermes: Human-to-robot embodied learning from multi-source motion data for mobile dexterous manipulation},
  author={Yuan, Zhecheng and Wei, Tianming and Gu, Langzhe and Hua, Pu and Liang, Tianhai and Chen, Yuanpei and Xu, Huazhe},
  journal={arXiv preprint arXiv:2508.20085},
  year={2025}
}

@inproceedings{kareer2025egomimic,
  title={Egomimic: Scaling imitation learning via egocentric video},
  author={Kareer, Simar and Patel, Dhruv and Punamiya, Ryan and Mathur, Pranay and Cheng, Shuo and Wang, Chen and Hoffman, Judy and Xu, Danfei},
  booktitle={2025 IEEE International Conference on Robotics and Automation (ICRA)},
  pages={13226--13233},
  year={2025},
  organization={IEEE}
}

@article{zhao2023learning,
  title={Learning fine-grained bimanual manipulation with low-cost hardware},
  author={Zhao, Tony Z and Kumar, Vikash and Levine, Sergey and Finn, Chelsea},
  journal={arXiv preprint arXiv:2304.13705},
  year={2023}
}

@article{zhou2023uni3d,
  title={Uni3d: Exploring unified 3d representation at scale},
  author={Zhou, Junsheng and Wang, Jinsheng and Ma, Baorui and Liu, Yu-Shen and Huang, Tiejun and Wang, Xinlong},
  journal={arXiv preprint arXiv:2310.06773},
  year={2023}
}

@article{beyer2024paligemma,
  title={Paligemma: A versatile 3b vlm for transfer},
  author={Beyer, Lucas and Steiner, Andreas and Pinto, Andr{\'e} Susano and Kolesnikov, Alexander and Wang, Xiao and Salz, Daniel and Neumann, Maxim and Alabdulmohsin, Ibrahim and Tschannen, Michael and Bugliarello, Emanuele and others},
  journal={arXiv preprint arXiv:2407.07726},
  year={2024}
}

@inproceedings{zhai2023sigmoid,
  title={Sigmoid loss for language image pre-training},
  author={Zhai, Xiaohua and Mustafa, Basil and Kolesnikov, Alexander and Beyer, Lucas},
  booktitle={Proceedings of the IEEE/CVF international conference on computer vision},
  pages={11975--11986},
  year={2023}
}

@article{ravi2024sam,
  title={Sam 2: Segment anything in images and videos},
  author={Ravi, Nikhila and Gabeur, Valentin and Hu, Yuan-Ting and Hu, Ronghang and Ryali, Chaitanya and Ma, Tengyu and Khedr, Haitham and R{\"a}dle, Roman and Rolland, Chloe and Gustafson, Laura and others},
  journal={arXiv preprint arXiv:2408.00714},
  year={2024}
}

@inproceedings{kwon2021h2o,
  title={H2o: Two hands manipulating objects for first person interaction recognition},
  author={Kwon, Taein and Tekin, Bugra and St{\"u}hmer, Jan and Bogo, Federica and Pollefeys, Marc},
  booktitle={Proceedings of the IEEE/CVF international conference on computer vision},
  pages={10138--10148},
  year={2021}
}

@inproceedings{liu2022hoi4d,
  title={Hoi4d: A 4d egocentric dataset for category-level human-object interaction},
  author={Liu, Yunze and Liu, Yun and Jiang, Che and Lyu, Kangbo and Wan, Weikang and Shen, Hao and Liang, Boqiang and Fu, Zhoujie and Wang, He and Yi, Li},
  booktitle={Proceedings of the IEEE/CVF Conference on Computer Vision and Pattern Recognition},
  pages={21013--21022},
  year={2022}
}

@article{banerjee2024hot3d,
  title={{HOT3D}: Hand and Object Tracking in {3D} from Egocentric Multi-View Videos},
  author={Banerjee, Prithviraj and Shkodrani, Sindi and Moulon, Pierre and Hampali, Shreyas and Han, Shangchen and Zhang, Fan and Zhang, Linguang and Fountain, Jade and Miller, Edward and Basol, Selen and Newcombe, Richard and Wang, Robert and Engel, Jakob Julian and Hodan, Tomas},
  journal={CVPR},
  year={2025}
}

@inproceedings{liu2024taco,
  title={Taco: Benchmarking generalizable bimanual tool-action-object understanding},
  author={Liu, Yun and Yang, Haolin and Si, Xu and Liu, Ling and Li, Zipeng and Zhang, Yuxiang and Liu, Yebin and Yi, Li},
  booktitle={Proceedings of the IEEE/CVF Conference on Computer Vision and Pattern Recognition},
  pages={21740--21751},
  year={2024}
}

@article{xue2025demogen,
  title={Demogen: Synthetic demonstration generation for data-efficient visuomotor policy learning},
  author={Xue, Zhengrong and Deng, Shuying and Chen, Zhenyang and Wang, Yixuan and Yuan, Zhecheng and Xu, Huazhe},
  journal={arXiv preprint arXiv:2502.16932},
  year={2025}
}

@article{zhong2025dexgraspvla,
  title={Dexgraspvla: A vision-language-action framework towards general dexterous grasping},
  author={Zhong, Yifan and Huang, Xuchuan and Li, Ruochong and Zhang, Ceyao and Chen, Zhang and Guan, Tianrui and Zeng, Fanlian and Lui, Ka Num and Ye, Yuyao and Liang, Yitao and others},
  journal={arXiv preprint arXiv:2502.20900},
  year={2025}
}

@article{wang2023mimicplay,
  title={Mimicplay: Long-horizon imitation learning by watching human play},
  author={Wang, Chen and Fan, Linxi and Sun, Jiankai and Zhang, Ruohan and Fei-Fei, Li and Xu, Danfei and Zhu, Yuke and Anandkumar, Anima},
  journal={arXiv preprint arXiv:2302.12422},
  year={2023}
}

@inproceedings{qin2022dexmv,
  title={Dexmv: Imitation learning for dexterous manipulation from human videos},
  author={Qin, Yuzhe and Wu, Yueh-Hua and Liu, Shaowei and Jiang, Hanwen and Yang, Ruihan and Fu, Yang and Wang, Xiaolong},
  booktitle={European Conference on Computer Vision},
  pages={570--587},
  year={2022},
  organization={Springer}
}

@article{ponce1997computing,
  title={On computing four-finger equilibrium and force-closure grasps of polyhedral objects},
  author={Ponce, Jean and Sullivan, Steve and Sudsang, Attawith and Boissonnat, Jean-Daniel and Merlet, Jean-Pierre},
  journal={The International Journal of Robotics Research},
  volume={16},
  number={1},
  pages={11--35},
  year={1997},
  publisher={Sage Publications Sage CA: Thousand Oaks, CA}
}

@article{kerr1986analysis,
  title={Analysis of multifingered hands},
  author={Kerr, Jeffrey and Roth, Bernard},
  journal={The International Journal of Robotics Research},
  volume={4},
  number={4},
  pages={3--17},
  year={1986},
  publisher={Sage Publications Sage CA: Thousand Oaks, CA}
}

@inproceedings{mordatch2012contact,
  title={Contact-invariant optimization for hand manipulation},
  author={Mordatch, Igor and Popovi{\'c}, Zoran and Todorov, Emanuel},
  booktitle={Proceedings of the ACM SIGGRAPH/Eurographics symposium on computer animation},
  pages={137--144},
  year={2012}
}

@inproceedings{bai2014dexterous,
  title={Dexterous manipulation using both palm and fingers},
  author={Bai, Yunfei and Liu, C Karen},
  booktitle={2014 IEEE International Conference on Robotics and Automation (ICRA)},
  pages={1560--1565},
  year={2014},
  organization={IEEE}
}

@inproceedings{li2025maniptrans,
  title={Maniptrans: Efficient dexterous bimanual manipulation transfer via residual learning},
  author={Li, Kailin and Li, Puhao and Liu, Tengyu and Li, Yuyang and Huang, Siyuan},
  booktitle={Proceedings of the Computer Vision and Pattern Recognition Conference},
  pages={6991--7003},
  year={2025}
}

@article{arimoto2004intelligent,
  title={Intelligent control of multi-fingered hands},
  author={Arimoto, Suguru},
  journal={Annual Reviews in Control},
  volume={28},
  number={1},
  pages={75--85},
  year={2004},
  publisher={Elsevier}
}

@article{chen2023visual,
  title={Visual dexterity: In-hand reorientation of novel and complex object shapes},
  author={Chen, Tao and Tippur, Megha and Wu, Siyang and Kumar, Vikash and Adelson, Edward and Agrawal, Pulkit},
  journal={Science Robotics},
  volume={8},
  number={84},
  pages={eadc9244},
  year={2023},
  publisher={American Association for the Advancement of Science}
}

@inproceedings{qi2023general,
  title={General in-hand object rotation with vision and touch},
  author={Qi, Haozhi and Yi, Brent and Suresh, Sudharshan and Lambeta, Mike and Ma, Yi and Calandra, Roberto and Malik, Jitendra},
  booktitle={Conference on Robot Learning},
  pages={2549--2564},
  year={2023},
  organization={PMLR}
}

@article{fang2025anydexgrasp,
  title={AnyDexGrasp: General Dexterous Grasping for Different Hands with Human-level Learning Efficiency},
  author={Fang, Hao-Shu and Yan, Hengxu and Tang, Zhenyu and Fang, Hongjie and Wang, Chenxi and Lu, Cewu},
  journal={arXiv preprint arXiv:2502.16420},
  year={2025}
}

@inproceedings{lin2025learning,
  title={Learning visuotactile skills with two multifingered hands},
  author={Lin, Toru and Zhang, Yu and Li, Qiyang and Qi, Haozhi and Yi, Brent and Levine, Sergey and Malik, Jitendra},
  booktitle={2025 IEEE International Conference on Robotics and Automation (ICRA)},
  pages={5637--5643},
  year={2025},
  organization={IEEE}
}

@article{akkaya2019solving,
  title={Solving rubik's cube with a robot hand},
  author={Akkaya, Ilge and Andrychowicz, Marcin and Chociej, Maciek and Litwin, Mateusz and McGrew, Bob and Petron, Arthur and Paino, Alex and Plappert, Matthias and Powell, Glenn and Ribas, Raphael and others},
  journal={arXiv preprint arXiv:1910.07113},
  year={2019}
}

@article{yin2023rotating,
  title={Rotating without seeing: Towards in-hand dexterity through touch},
  author={Yin, Zhao-Heng and Huang, Binghao and Qin, Yuzhe and Chen, Qifeng and Wang, Xiaolong},
  journal={arXiv preprint arXiv:2303.10880},
  year={2023}
}

@article{wang2024dexcap,
  title={Dexcap: Scalable and portable mocap data collection system for dexterous manipulation},
  author={Wang, Chen and Shi, Haochen and Wang, Weizhuo and Zhang, Ruohan and Fei-Fei, Li and Liu, C Karen},
  journal={arXiv preprint arXiv:2403.07788},
  year={2024}
}

@article{he2025dexvlg,
  title={DexVLG: Dexterous Vision-Language-Grasp Model at Scale},
  author={He, Jiawei and Li, Danshi and Yu, Xinqiang and Qi, Zekun and Zhang, Wenyao and Chen, Jiayi and Zhang, Zhaoxiang and Zhang, Zhizheng and Yi, Li and Wang, He},
  journal={arXiv preprint arXiv:2507.02747},
  year={2025}
}

@article{cheng2024open,
  title={Open-television: Teleoperation with immersive active visual feedback},
  author={Cheng, Xuxin and Li, Jialong and Yang, Shiqi and Yang, Ge and Wang, Xiaolong},
  journal={arXiv preprint arXiv:2407.01512},
  year={2024}
}

@inproceedings{qin2023anyteleop,
  title     = {AnyTeleop: A General Vision-Based Dexterous Robot Arm-Hand Teleoperation System},
  author    = {Qin, Yuzhe and Yang, Wei and Huang, Binghao and Van Wyk, Karl and Su, Hao and Wang, Xiaolong and Chao, Yu-Wei and Fox, Dieter},
  booktitle = {Robotics: Science and Systems},
  year      = {2023}
}

@article{dalal2023imitating,
  title={Imitating task and motion planning with visuomotor transformers},
  author={Dalal, Murtaza and Mandlekar, Ajay and Garrett, Caelan and Handa, Ankur and Salakhutdinov, Ruslan and Fox, Dieter},
  journal={arXiv preprint arXiv:2305.16309},
  year={2023}
}

@article{yuan2025motiontrans,
  title={Motiontrans: Human vr data enable motion-level learning for robotic manipulation policies},
  author={Yuan, Chengbo and Zhou, Rui and Liu, Mengzhen and Hu, Yingdong and Wang, Shengjie and Yi, Li and Wen, Chuan and Zhang, Shanghang and Gao, Yang},
  journal={arXiv preprint arXiv:2509.17759},
  year={2025}
}

@article{yuan2024general,
  title={General flow as foundation affordance for scalable robot learning},
  author={Yuan, Chengbo and Wen, Chuan and Zhang, Tong and Gao, Yang},
  journal={arXiv preprint arXiv:2401.11439},
  year={2024}
}

@article{wen2023any,
  title={Any-point trajectory modeling for policy learning},
  author={Wen, Chuan and Lin, Xingyu and So, John and Chen, Kai and Dou, Qi and Gao, Yang and Abbeel, Pieter},
  journal={arXiv preprint arXiv:2401.00025},
  year={2023}
}

@article{chen2024object,
  title={Object-centric dexterous manipulation from human motion data},
  author={Chen, Yuanpei and Wang, Chen and Yang, Yaodong and Liu, C Karen},
  journal={arXiv preprint arXiv:2411.04005},
  year={2024}
}

@article{ye2024latent,
  title={Latent action pretraining from videos},
  author={Ye, Seonghyeon and Jang, Joel and Jeon, Byeongguk and Joo, Sejune and Yang, Jianwei and Peng, Baolin and Mandlekar, Ajay and Tan, Reuben and Chao, Yu-Wei and Lin, Bill Yuchen and others},
  journal={arXiv preprint arXiv:2410.11758},
  year={2024}
}

@article{niu2025pre,
  title={Pre-training auto-regressive robotic models with 4d representations},
  author={Niu, Dantong and Sharma, Yuvan and Xue, Haoru and Biamby, Giscard and Zhang, Junyi and Ji, Ziteng and Darrell, Trevor and Herzig, Roei},
  journal={arXiv preprint arXiv:2502.13142},
  year={2025}
}

@article{nair2022r3m,
  title={R3m: A universal visual representation for robot manipulation},
  author={Nair, Suraj and Rajeswaran, Aravind and Kumar, Vikash and Finn, Chelsea and Gupta, Abhinav},
  journal={arXiv preprint arXiv:2203.12601},
  year={2022}
}

@article{zeng2024learning,
  title={Learning manipulation by predicting interaction},
  author={Zeng, Jia and Bu, Qingwen and Wang, Bangjun and Xia, Wenke and Chen, Li and Dong, Hao and Song, Haoming and Wang, Dong and Hu, Di and Luo, Ping and others},
  journal={arXiv preprint arXiv:2406.00439},
  year={2024}
}

@article{liu2025egozero,
  title={Egozero: Robot learning from smart glasses},
  author={Liu, Vincent and Adeniji, Ademi and Zhan, Haotian and Haldar, Siddhant and Bhirangi, Raunaq and Abbeel, Pieter and Pinto, Lerrel},
  journal={arXiv preprint arXiv:2505.20290},
  year={2025}
}

@article{fourier2025actionnet,
  author    = {Fourier ActionNet Team, Yao Mu},
  title     = {ActionNet: A dataset for dexterous bimanual manipulation},
  year      = {2025},
}

@article{wu2024robomind,
  title={Robomind: Benchmark on multi-embodiment intelligence normative data for robot manipulation},
  author={Wu, Kun and Hou, Chengkai and Liu, Jiaming and Che, Zhengping and Ju, Xiaozhu and Yang, Zhuqin and Li, Meng and Zhao, Yinuo and Xu, Zhiyuan and Yang, Guang and others},
  journal={arXiv preprint arXiv:2412.13877},
  year={2024}
}

@article{liu2024realdex,
  title={Realdex: Towards human-like grasping for robotic dexterous hand},
  author={Liu, Yumeng and Yang, Yaxun and Wang, Youzhuo and Wu, Xiaofei and Wang, Jiamin and Yao, Yichen and Schwertfeger, S{\"o}ren and Yang, Sibei and Wang, Wenping and Yu, Jingyi and others},
  journal={arXiv preprint arXiv:2402.13853},
  year={2024}
}

@book{hartley2003multiple,
  title={Multiple view geometry in computer vision},
  author={Hartley, Richard},
  volume={665},
  year={2003},
  publisher={Cambridge university press}
}

@misc{coumans2016pybullet,
  title={Pybullet, a python module for physics simulation for games, robotics and machine learning},
  author={Coumans, Erwin and Bai, Yunfei},
  year={2016}
}

@article{dosovitskiy2020image,
  title={An image is worth 16x16 words: Transformers for image recognition at scale},
  author={Dosovitskiy, Alexey},
  journal={arXiv preprint arXiv:2010.11929},
  year={2020}
}

@article{lipman2022flow,
  title={Flow matching for generative modeling},
  author={Lipman, Yaron and Chen, Ricky TQ and Ben-Hamu, Heli and Nickel, Maximilian and Le, Matt},
  journal={arXiv preprint arXiv:2210.02747},
  year={2022}
}

@article{liu2024rdt,
  title={Rdt-1b: a diffusion foundation model for bimanual manipulation},
  author={Liu, Songming and Wu, Lingxuan and Li, Bangguo and Tan, Hengkai and Chen, Huayu and Wang, Zhengyi and Xu, Ke and Su, Hang and Zhu, Jun},
  journal={arXiv preprint arXiv:2410.07864},
  year={2024}
}

@article{zhang2025align,
  title={Align-Then-stEer: Adapting the Vision-Language Action Models through Unified Latent Guidance},
  author={Zhang, Yang and Wang, Chenwei and Lu, Ouyang and Zhao, Yuan and Ge, Yunfei and Sun, Zhenglong and Li, Xiu and Zhang, Chi and Bai, Chenjia and Li, Xuelong},
  journal={arXiv preprint arXiv:2509.02055},
  year={2025}
}

@article{bu2025univla,
  title={Univla: Learning to act anywhere with task-centric latent actions},
  author={Bu, Qingwen and Yang, Yanting and Cai, Jisong and Gao, Shenyuan and Ren, Guanghui and Yao, Maoqing and Luo, Ping and Li, Hongyang},
  journal={arXiv preprint arXiv:2505.06111},
  year={2025}
}

@inproceedings{potamias2025wilor,
  title={Wilor: End-to-end 3d hand localization and reconstruction in-the-wild},
  author={Potamias, Rolandos Alexandros and Zhang, Jinglei and Deng, Jiankang and Zafeiriou, Stefanos},
  booktitle={Proceedings of the Computer Vision and Pattern Recognition Conference},
  pages={12242--12254},
  year={2025}
}

@inproceedings{zhong2025dexgrasp,
  title={Dexgrasp anything: Towards universal robotic dexterous grasping with physics awareness},
  author={Zhong, Yiming and Jiang, Qi and Yu, Jingyi and Ma, Yuexin},
  booktitle={Proceedings of the Computer Vision and Pattern Recognition Conference},
  pages={22584--22594},
  year={2025}
}

@inproceedings{tian2025pdfactor,
  title={PDFactor: Learning Tri-Perspective View Policy Diffusion Field for Multi-Task Robotic Manipulation},
  author={Tian, Jingyi and Wang, Le and Zhou, Sanping and Wang, Sen and Li, Jiayi and Sun, Haowen and Tang, Wei},
  booktitle={Proceedings of the Computer Vision and Pattern Recognition Conference},
  pages={15757--15767},
  year={2025}
}

@inproceedings{zhou2025mitigating,
  title={Mitigating the human-robot domain discrepancy in visual pre-training for robotic manipulation},
  author={Zhou, Jiaming and Ma, Teli and Lin, Kun-Yu and Wang, Zifan and Qiu, Ronghe and Liang, Junwei},
  booktitle={Proceedings of the Computer Vision and Pattern Recognition Conference},
  pages={22551--22561},
  year={2025}
}

@article{jian2025g,
  title={G-DexGrasp: Generalizable Dexterous Grasping Synthesis Via Part-Aware Prior Retrieval and Prior-Assisted Generation},
  author={Jian, Juntao and Liu, Xiuping and Chen, Zixuan and Li, Manyi and Liu, Jian and Hu, Ruizhen},
  journal={arXiv preprint arXiv:2503.19457},
  year={2025}
}

@article{wang2025dexh2r,
  title={DexH2R: A Benchmark for Dynamic Dexterous Grasping in Human-to-Robot Handover},
  author={Wang, Youzhuo and Ye, Jiayi and Xiao, Chuyang and Zhong, Yiming and Tao, Heng and Yu, Hang and Liu, Yumeng and Yu, Jingyi and Ma, Yuexin},
  journal={arXiv preprint arXiv:2506.23152},
  year={2025}
}

@inproceedings{miao2025fedvla,
  title={Fedvla: Federated vision-language-action learning with dual gating mixture-of-experts for robotic manipulation},
  author={Miao, Cui and Chang, Tao and Wu, Meihan and Xu, Hongbin and Li, Chun and Li, Ming and Wang, Xiaodong},
  booktitle={Proceedings of the IEEE/CVF International Conference on Computer Vision},
  pages={6904--6913},
  year={2025}
}

@inproceedings{li2025learning,
  title={Learning precise affordances from egocentric videos for robotic manipulation},
  author={Li, Gen and Tsagkas, Nikolaos and Song, Jifei and Mon-Williams, Ruaridh and Vijayakumar, Sethu and Shao, Kun and Sevilla-Lara, Laura},
  booktitle={Proceedings of the IEEE/CVF International Conference on Computer Vision},
  pages={10581--10591},
  year={2025}
}

@inproceedings{ji2025robobrain,
  title={Robobrain: A unified brain model for robotic manipulation from abstract to concrete},
  author={Ji, Yuheng and Tan, Huajie and Shi, Jiayu and Hao, Xiaoshuai and Zhang, Yuan and Zhang, Hengyuan and Wang, Pengwei and Zhao, Mengdi and Mu, Yao and An, Pengju and others},
  booktitle={Proceedings of the Computer Vision and Pattern Recognition Conference},
  pages={1724--1734},
  year={2025}
}

@inproceedings{chen2025vidbot,
  title={VidBot: Learning Generalizable 3D Actions from In-the-Wild 2D Human Videos for Zero-Shot Robotic Manipulation},
  author={Chen, Hanzhi and Sun, Boyang and Zhang, Anran and Pollefeys, Marc and Leutenegger, Stefan},
  booktitle={Proceedings of the Computer Vision and Pattern Recognition Conference},
  pages={27661--27672},
  year={2025}
}

@inproceedings{xu2025diffusion,
  title={Diffusion-Based Imaginative Coordination for Bimanual Manipulation},
  author={Xu, Huilin and Ding, Jian and Xu, Jiakun and Wang, Ruixiang and Chen, Jun and Mai, Jinjie and Fu, Yanwei and Ghanem, Bernard and Xu, Feng and Elhoseiny, Mohamed},
  booktitle={Proceedings of the IEEE/CVF International Conference on Computer Vision},
  pages={11469--11479},
  year={2025}
}

@inproceedings{zhang2025elucidating,
  title={Elucidating the Design Space of Torque-aware Vision-Language-Action Models},
  author={Zhang, Zongzheng and Xu, Haobo and Yang, Zhuo and Yue, Chenghao and Lin, Zehao and Gao, Huan-ang and Wang, Ziwei and Zhao, Hao},
  booktitle={9th Annual Conference on Robot Learning},
  year={2025}
}

@article{zhang2025robochemist,
  title={RoboChemist: Long-Horizon and Safety-Compliant Robotic Chemical Experimentation},
  author={Zhang, Zongzheng and Yue, Chenghao and Xu, Haobo and Liao, Minwen and Qi, Xianglin and Gao, Huan-ang and Wang, Ziwei and Zhao, Hao},
  journal={arXiv preprint arXiv:2509.08820},
  year={2025}
}

@inproceedings{ding2024preafford,
  title={Preafford: Universal affordance-based pre-grasping for diverse objects and environments},
  author={Ding, Kairui and Chen, Boyuan and Wu, Ruihai and Li, Yuyang and Zhang, Zongzheng and Gao, Huan-ang and Li, Siqi and Zhou, Guyue and Zhu, Yixin and Dong, Hao and others},
  booktitle={2024 IEEE/RSJ International Conference on Intelligent Robots and Systems (IROS)},
  pages={7278--7285},
  year={2024},
  organization={IEEE}
}

@inproceedings{zhang2023flexible,
  title={Flexible handover with real-time robust dynamic grasp trajectory generation},
  author={Zhang, Gu and Fang, Hao-Shu and Fang, Hongjie and Lu, Cewu},
  booktitle={2023 IEEE/RSJ International Conference on Intelligent Robots and Systems (IROS)},
  pages={3192--3199},
  year={2023},
  organization={IEEE}
}

@article{si2024difftactile,
  title={Difftactile: A physics-based differentiable tactile simulator for contact-rich robotic manipulation},
  author={Si, Zilin and Zhang, Gu and Ben, Qingwei and Romero, Branden and Xian, Zhou and Liu, Chao and Gan, Chuang},
  journal={arXiv preprint arXiv:2403.08716},
  year={2024}
}

@article{yuan2024learning,
  title={Learning to manipulate anywhere: A visual generalizable framework for reinforcement learning},
  author={Yuan, Zhecheng and Wei, Tianming and Cheng, Shuiqi and Zhang, Gu and Chen, Yuanpei and Xu, Huazhe},
  journal={arXiv preprint arXiv:2407.15815},
  year={2024}
}
}

% \clearpage
% \setcounter{page}{1}
% \maketitlesupplementary

\clearpage
\setcounter{page}{1}

\onecolumn 

% ===== Big Appendix title =====
\begin{center}
    {\LARGE\bfseries Appendix \\[0.5em]}
\end{center}
\vspace{1em}

\appendix

% \section{Video Visualization}
% As an attachment, we provide an \texttt{.mp4} file containing video visualizations of a subset of UniDex-Dataset as well as the real-world experiment rollouts.

% \section{More Visualization of UniDex-Dataset}
% We present additional visualizations from a subset of UniDex-Dataset in Fig.~\ref{fig:morevis}. For several representative tasks, we display temporally consecutive frames after human-in-the-loop retargeting, highlighting the temporally consistent hand–object interactions captured in UniDex-Dataset.

\section{Training Details}
\label{sec:ap-training-detail}
\subsection{UniDex-VLA Flow-Matching Loss}

To train UniDex-VLA, we minimize a conditional flow-matching loss:
\begin{equation}
L^{\tau}(\theta)
= \mathbb{E}_{p(A_t \mid o_t),\, q(A_t^{\tau} \mid A_t)}\left[
\bigl\Vert v_{\theta}(A_t^{\tau}, o_t) - u(A_t^{\tau} \mid A_t) \bigr\Vert\right],
\end{equation}
where $\tau \in [0,1]$ and $q(A_t^{\tau} \mid A_t)=\mathcal{N}\!\bigl(\tau A_t,\; (1-\tau)I\bigr)$ is a linear-Gaussian probability path. We sample
$A_t^{\tau} = \tau A_t + (1-\tau)\epsilon$ with $\epsilon \sim \mathcal{N}(0,I)$ and compute the target conditional vector field
$u(A_t^{\tau} \mid A_t) = A_t - \epsilon$. The network is trained such that the predicted vector field $v_{\theta}(A_t^{\tau}, o_t)$ approximates $u(A_t^{\tau} \mid A_t)$.

At inference time, we integrate the learned vector field using a forward Euler scheme to generate a denoised action chunk:
\[
A_t^{\tau+\delta} \;=\; A_t^{\tau} + \delta\, v_{\theta}\!\left(A_t^{\tau},\, o_t\right),
\]
with step size $\delta = 0.1$ and initial condition $A_t^{0} \sim \mathcal{N}(0,I)$.

\subsection{UniDex-VLA Pretraining}

During pre-training, we use 8 NVIDIA H800 GPUs with a total batch size of 128. The model used for subsequent post-training is trained for 3 epochs ($\sim$ 30k steps), which takes around 24 hours. We adopt the AdamW optimizer and a cosine learning-rate scheduler with an initial learning rate of 1e-4. The learning rate is decayed by a factor of 0.95 at the 2nd epoch. The weight decay is set to 1e-10, and we apply gradient clipping with a maximum norm of 1.0.

\subsection{UniDex-VLA Post-training}

During post-training, we use 2 NVIDIA H800 GPUs for each task, with a total batch size of 8. We use the AdamW optimizer without a learning rate scheduler and set the initial learning rate to 2.5e-5. For common data, we train the model for 50 epochs ($\sim$ 3k steps), which takes around 4 hours. For DemoGen~\cite{xue2025demogen} augmented data, we train the model for 2 epochs ($\sim$ 1.8k steps), which takes around 2.5 hours. The weight decay is set to 1e-10, and we again use gradient clipping with a maximum norm of 1.0.

% Specifically, the teleoperation control frequency is 15 fps; we downsample the collected data to 5 fps for post-training and use the same 5 fps control frequency during inference.

\subsection{Baselines}

For all baselines (DP~\cite{dp}, DP3~\cite{dp3}, and $\pi_0$~\cite{black2024pi_0}), we post-train the models until convergence on the validation set.

For DP~\cite{dp} and DP3~\cite{dp3}, we use the AdamW optimizer with an initial learning rate of 1e-4. The state horizon is set to 4 and the action horizon to 32. We use a batch size of 32 and train for 400 epochs. 

For $\pi_0$~\cite{black2024pi_0}, we use the AdamW optimizer with an initial learning rate of 2.5e-5. The batch size is set to 8 and the model is trained for 50 epochs. The number of diffusion steps is set to 10. 

For our UniDex-VLA baseline without pretraining, we use the same training hyperparameters      as UniDex-VLA with pretraining.

\section{Human-in-the-loop Retargeting GUI}

\begin{figure*}[h]
\centering
  \includegraphics[width=1.0\textwidth]{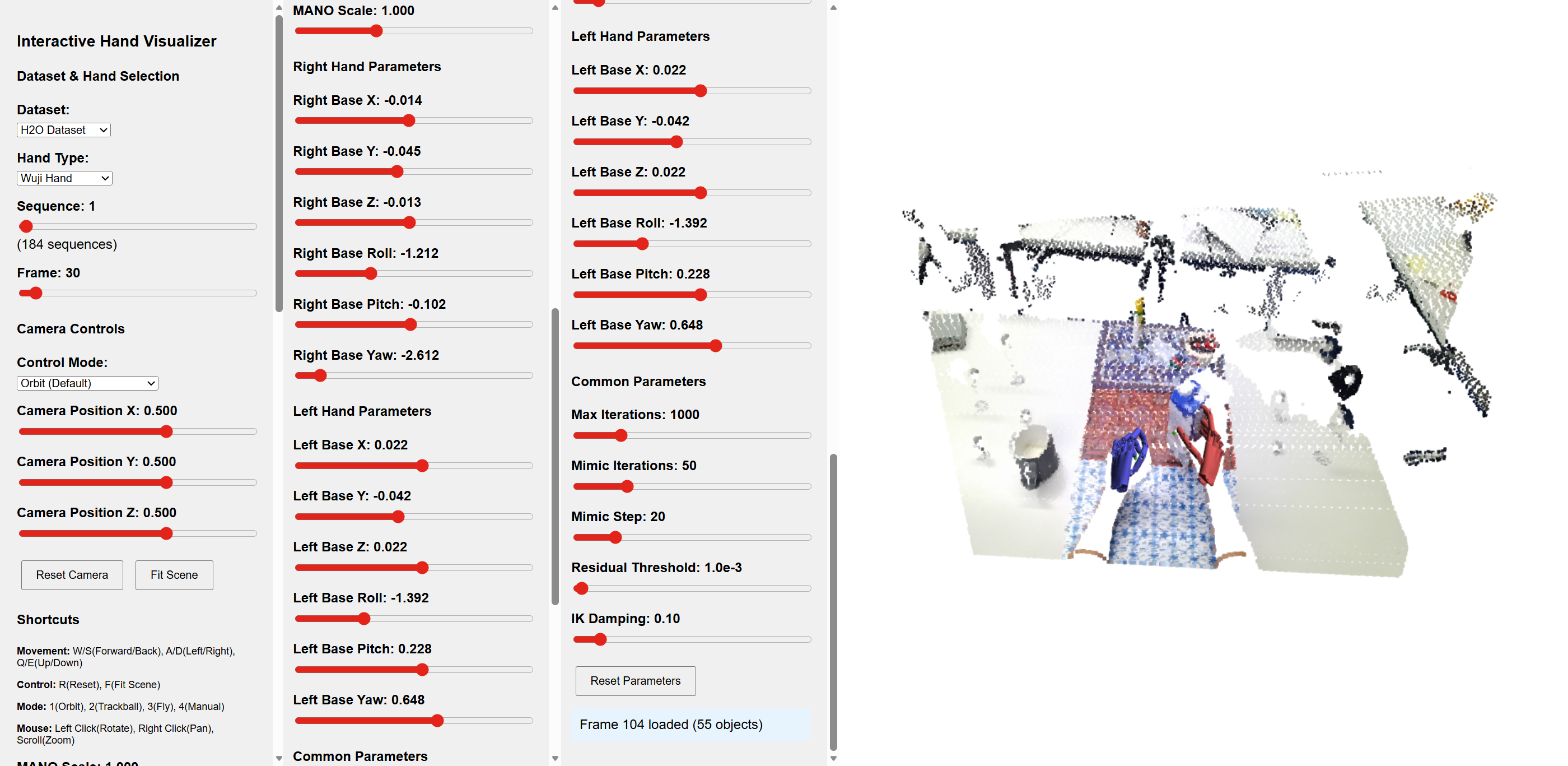}
  \caption{Human-friendly web-basedGUI for retargeting human demonstrations to robot executions. Users can adjust the IK parameters, dummy links, and other settings through the GUI to obtain satisfactory retargeted robot trajectories.}
  \label{fig:web_gui}
 \vspace*{-10pt}
\end{figure*}

To minimum human efforts in our human-in-the-loop retargeting process, we develop a human-friendly web-based GUI, as shown in \cref{fig:web_gui}. Through this interface, users can adjust dummy base links, IK parameters, and other retargeting settings to obtain satisfactory robot trajectories. 

\section{FAAS Details}
\label{sec:ap-faas}

\begin{figure*}[h]
\centering
  \includegraphics[width=1.0\textwidth]{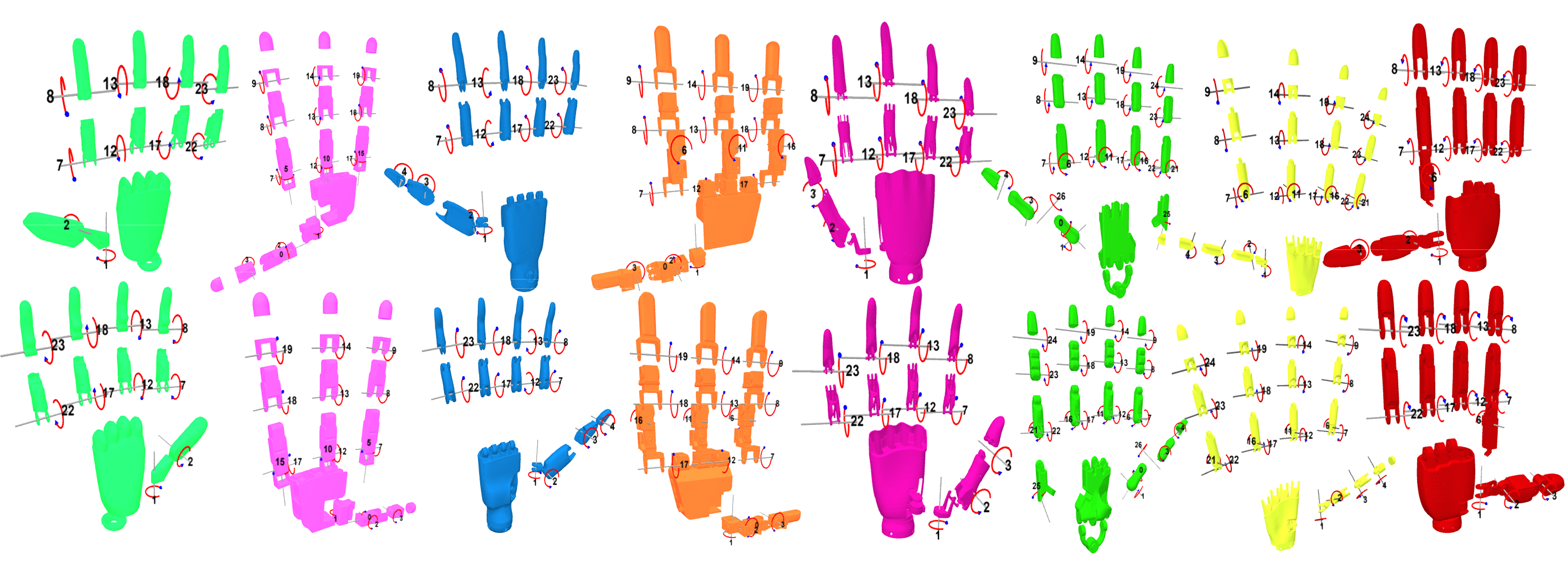}
  \caption{Joint mappings of different robotic hands used in FAAS. From left to right are Ability, Allegro, Inspire, Leap, Oymotion, Shadow, Wuji, and Xhand. The two rows show different views of the joint mappings on the right hand.}
  \label{fig:detailed_faas}
 \vspace*{-10pt}
\end{figure*}

Here we show the details for the 32 dimensions encoding dexterous hand joints. Dimensions 0–4, 5–9, 10–14, 15–19, and 20–24 correspond to the thumb, index, middle, ring, and little fingers, respectively. Dimensions 25–26 are reserved for extra wrist joints of Shadow hands. Dimensions 27–31 are left unused for new hands. The detailed joint mappings of the robotic hands used in FAAS are shown in \cref{fig:detailed_faas}.

\section{UniDex-Cap Setup Calibration}

UniDex-Cap combines an Apple Vision Pro (for hand and head poses, denoted $\{P_{\mathrm{VP}}\}$) and an Intel RealSense L515 (for RGB-D). Because Vision Pro does not expose third-party RGB-D video recording, we physically couple the two sensors with a custom 3D-printed mount that rigidly fixes their relative pose. This mechanical constraint ensures that the extrinsic transform between the Vision Pro and the RealSense remains stable for a given user.

\begin{figure*}[h]

\centering
  \includegraphics[width=1.0\textwidth]{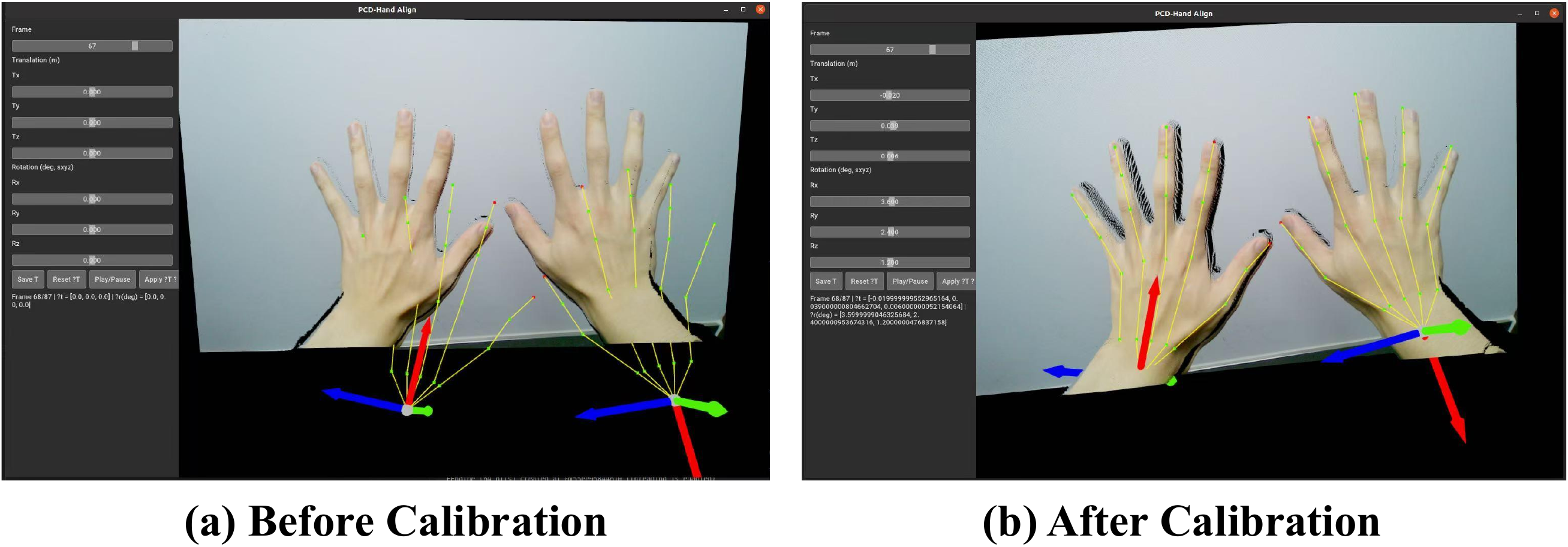}
  \caption{GUI for UniDex-Cap calibration. (a) shows the initial state before calibration; (b) shows the calibrated result where the hand poses captured by Vision Pro align with the 3D point cloud captured by the RealSense L515 camera.}
  \label{fig:calibration}
 \vspace*{-15pt}
\end{figure*}

As shown in Fig.~\ref{fig:calibration}, we provide a lightweight GUI to estimate the remaining constant extrinsics with minimal manual effort. The user records a short calibration clip and then uses a slider-based interface to adjust the hand and wrist poses in the Vision Pro coordinate frame—visualized as a skeleton—until they align with the 3D hand point cloud captured by the RealSense camera. The slider values directly correspond to the transform $T^{\mathrm{VP}}_{\mathrm{RS}}$. Once this transform is determined, all Vision Pro poses are converted into the RealSense camera frame, yielding temporally aligned hand and head trajectories:
\begin{equation}
    P_{\mathrm{RS}} \;=\; T^{\mathrm{VP}}_{\mathrm{RS}} \, P_{\mathrm{VP}},
\end{equation}
where $P_{\mathrm{VP}}$ and $P_{\mathrm{RS}}$ are represented in homogeneous coordinates. This pipeline produces temporally synchronized, geometrically consistent annotations suitable for downstream retargeting and post-training.

\section{Core Contribution List}
\label{sec:contrib}
The main contributions of the core contributors are as follows:

\noindent\textbf{Gu Zhang}: Project lead. Developed the overall dataset construction pipeline, model architecture, and unified action space; built the robot system infrastructure; and wrote the paper.

\noindent\textbf{Qicheng Xu}: Led VLA model training; optimized the dataset construction and policy inference pipelines; and contributed to paper writing.

\noindent\textbf{Haozhe Zhang}: Led dataset processing; improved the robot system and the human--robot data capture pipeline; and contributed to paper writing.

\noindent\textbf{Jianhan Ma}: Implemented retargeting algorithms; and developed dataset visualizations and contributed to early-stage exploration.

\noindent\textbf{Long He}: Implemented DemoGen algorithm; and contributed to paper writing. 

\noindent\textbf{Yiming Bao}: Collected robot data and human data; and contributed to DemoGen algorithm implementation.

\end{document}